\pdfoutput=1

\documentclass[11pt]{article}

\usepackage[final]{acl}

\usepackage{times}
\usepackage{latexsym}

\usepackage[T1]{fontenc}

\usepackage[utf8]{inputenc}

\usepackage{microtype}

\usepackage{inconsolata}

\usepackage{graphicx}

\usepackage{makecell}
\usepackage{microtype}
\usepackage{hyperref}
\usepackage{url}
\usepackage{multicol} 
\usepackage{booktabs}
\usepackage{multirow}%
\usepackage{amsmath,amssymb,amsfonts}%
\usepackage{amsthm}%
\usepackage{mathrsfs}%
\usepackage[title]{appendix}%
\usepackage[normalem]{ulem}
\usepackage{xcolor}%
\usepackage{xspace}
\usepackage{textcomp}%
\usepackage{manyfoot}%
\usepackage{booktabs}%
\usepackage{algorithm}%
\usepackage{algorithmicx}%
\usepackage{algpseudocode}%
\usepackage{listings}%
\usepackage{subcaption}
\usepackage{commath}
\usepackage{tabularray}
\usepackage{bbding}
\usepackage{pifont}
\usepackage{utfsym}
\usepackage{times}
\usepackage{wasysym}
\usepackage{fontawesome}
\usepackage[normalem]{ulem}
\usepackage{colortbl}
\usepackage{lscape}
\usepackage{enumitem}
\usepackage{fontawesome}

\definecolor{myred}{HTML}{C00000}  
\definecolor{mygreen}{HTML}{548235}  

\newcommand{\agentbenchname}{\textsc{CAB}\xspace}
\newcommand{\agentname}{\textsc{ReflecTool}\xspace}

\title{\agentname: Towards Reflection-Aware Tool-Augmented Clinical Agents}

\author{Yusheng Liao$^{\spadesuit,\diamondsuit}$, 
Shuyang Jiang$^{\clubsuit,\diamondsuit}$,
Yanfeng Wang$^{\spadesuit,\diamondsuit}$,
Yu~Wang\thanks{Corresponding Author}$^{,\spadesuit,\diamondsuit}$
 \\
  $^{\spadesuit}$Shanghai Jiao Tong University \\
  $^{\diamondsuit}$Shanghai Artificial Intelligence Laboratory \\
  $^{\clubsuit}$Fudan University \\
  \texttt{\{liao20160907,wangyanfeng622,yuwangsjtu\}@sjtu.edu.cn} \\
  \texttt{shuyangjiang23@m.fudan.edu.cn}
}

\begin{document}

\maketitle
\begin{abstract}
Large Language Models (LLMs) have shown promising potential in the medical domain, assisting with tasks like clinical note generation and patient communication. 
However, current LLMs are limited to text-based communication, hindering their ability to interact with diverse forms of information in clinical environments. 
Despite clinical agents succeeding in diverse signal interaction, they are oriented to a single clinical scenario and hence fail for broader applications.
To evaluate clinical agents holistically, we propose ClinicalAgent Bench~(\agentbenchname), a comprehensive medical agent benchmark consisting of 18 tasks across five key realistic clinical dimensions. 
Building on this, we introduce \agentname, a novel framework that excels at utilizing domain-specific tools within two stages.
The first optimization stage progressively enlarges a long-term memory by saving successful solving processes and tool-wise experience of agents in a tiny pre-defined training set.
In the following inference stage, \agentname can search for supportive successful demonstrations from already built long-term memory to guide the tool selection strategy, 
and a verifier improves the tool usage according to the tool-wise experience with two verification methods--iterative refinement and candidate selection. 
Extensive experiments on \agentbenchname demonstrate that \agentname surpasses the pure LLMs with more than 10 points and the well-established agent-based methods with 3 points, highlighting its adaptability and effectiveness in solving complex clinical tasks. Our code and datasets are available at \url{https://github.com/BlueZeros/ReflecTool}.
\end{abstract}

\begin{table*}[]
\centering
\resizebox{\textwidth}{!}{%
\begin{tabular}{lccccccccc}
\toprule
\multirow{2}{*}{\textbf{Methods}} & \multicolumn{5}{c}{\textbf{Agent Capacities}} & \multicolumn{3}{c}{\textbf{Agent Methods}} &  \\
\cmidrule(lr){2-6}\cmidrule(lr){7-10}
 & \textbf{\begin{tabular}[c]{@{}c@{}}Knowledge\&\\ Reasoning\end{tabular}} & \textbf{MultiModal} & \textbf{\begin{tabular}[c]{@{}c@{}}Numerical\\ Analysis\end{tabular}} & \textbf{\begin{tabular}[c]{@{}c@{}}Data \\ Understanding\end{tabular}} & \textbf{\begin{tabular}[c]{@{}c@{}}Trustworthi\\ ness\end{tabular}} & \textbf{\begin{tabular}[c]{@{}c@{}}Tool\\ Use\end{tabular}} & \textbf{\begin{tabular}[c]{@{}c@{}}Long-Term\\ Memory\end{tabular}} & \textbf{\begin{tabular}[c]{@{}c@{}}Tool-wise\\ Reflection\end{tabular}} & \textbf{\begin{tabular}[c]{@{}c@{}}w/o Fine\\ Tuning\end{tabular}} \\
 \midrule
MedAgent~\citep{tang2024medagents} & \textcolor{mygreen}{\ding{51}} & \textcolor{myred}{\ding{55}} & \textcolor{myred}{\ding{55}} & \textcolor{mygreen}{\ding{51}} & \textcolor{myred}{\ding{55}} & \textcolor{myred}{\ding{55}} & \textcolor{myred}{\ding{55}} & \textcolor{myred}{\ding{55}} & \textcolor{mygreen}{\ding{51}} \\
MMedAgent~\citep{li2024mmedagent} & \textcolor{mygreen}{\ding{51}} & \textcolor{mygreen}{\ding{51}} & \textcolor{myred}{\ding{55}} & \textcolor{myred}{\ding{55}} & \textcolor{myred}{\ding{55}} & \textcolor{mygreen}{\ding{51}} & \textcolor{myred}{\ding{55}} & \textcolor{myred}{\ding{55}} & \textcolor{myred}{\ding{55}} \\
MedRAG~\citep{xiong-etal-2024-benchmarking} & \textcolor{mygreen}{\ding{51}} & \textcolor{myred}{\ding{55}} & \textcolor{myred}{\ding{55}} & \textcolor{myred}{\ding{55}} & \textcolor{myred}{\ding{55}} & \textcolor{mygreen}{\ding{51}} & \textcolor{myred}{\ding{55}} & \textcolor{myred}{\ding{55}} & \textcolor{mygreen}{\ding{51}} \\
OmniRAG~\citep{chen2025towards} & \textcolor{mygreen}{\ding{51}} & \textcolor{myred}{\ding{55}} & \textcolor{myred}{\ding{55}} & \textcolor{myred}{\ding{55}} & \textcolor{myred}{\ding{55}} & \textcolor{mygreen}{\ding{51}} & \textcolor{myred}{\ding{55}} & \textcolor{myred}{\ding{55}} & \textcolor{myred}{\ding{55}} \\
AgentMD~\citep{jin2024agentmd} & \textcolor{mygreen}{\ding{51}} & \textcolor{myred}{\ding{55}} & \textcolor{mygreen}{\ding{51}} & \textcolor{myred}{\ding{55}} & \textcolor{myred}{\ding{55}} & \textcolor{mygreen}{\ding{51}} & \textcolor{myred}{\ding{55}} & \textcolor{myred}{\ding{55}} & \textcolor{mygreen}{\ding{51}} \\
EHRAgent~\citep{shi2024ehragent} & \textcolor{mygreen}{\ding{51}} & \textcolor{myred}{\ding{55}} & \textcolor{mygreen}{\ding{51}} & \textcolor{myred}{\ding{55}} & \textcolor{mygreen}{\ding{51}} & \textcolor{mygreen}{\ding{51}} & \textcolor{mygreen}{\ding{51}} & \textcolor{myred}{\ding{55}} & \textcolor{mygreen}{\ding{51}} \\
CTAgent~\citep{yue2024ct} & \textcolor{mygreen}{\ding{51}} & \textcolor{mygreen}{\ding{51}} & \textcolor{myred}{\ding{55}} & \textcolor{myred}{\ding{55}} & \textcolor{mygreen}{\ding{51}} & \textcolor{mygreen}{\ding{51}} & \textcolor{myred}{\ding{55}} & \textcolor{myred}{\ding{55}} & \textcolor{myred}{\ding{55}} \\
BKGAgent~\citep{lin2024biokgbench} & \textcolor{mygreen}{\ding{51}} & \textcolor{myred}{\ding{55}} & \textcolor{myred}{\ding{55}} & \textcolor{myred}{\ding{55}} & \textcolor{myred}{\ding{55}} & \textcolor{mygreen}{\ding{51}} & \textcolor{myred}{\ding{55}} & \textcolor{myred}{\ding{55}} & \textcolor{mygreen}{\ding{51}} \\
\midrule
\agentname (Ours) & \textcolor{mygreen}{\ding{51}} & \textcolor{mygreen}{\ding{51}} & \textcolor{mygreen}{\ding{51}} & \textcolor{mygreen}{\ding{51}} & \textcolor{mygreen}{\ding{51}} & \textcolor{mygreen}{\ding{51}} & \textcolor{mygreen}{\ding{51}} & \textcolor{mygreen}{\ding{51}} & \textcolor{mygreen}{\ding{51}} \\
\bottomrule
\end{tabular}%
}
\caption{Comparison of previous medical agents and \agentname on both agent capacities and methods.}
\label{tab: comparision}
\end{table*}

\section{Introduction}

Large Language Models (LLMs) have shown significant potential in the medical domain~\citep{singhal2023large, nori2023capabilities, chen2023meditron}, demonstrating their ability to assist with tasks such as generating clinical notes~\citep{biswas2024intelligent,jung2024enhancing} and supporting patient communication~\citep{tu2024towards,liao2024automatic}. However, LLMs are restricted to direct text-based responses rather than serving as a bridge to leverage the information in other forms, thus impeding their effective application in realistic clinical scenarios.

To address such shortcoming, numerous works developed more advanced 
clinical agents, which enable models to leverage complex information through specialized tools~\citep{jin2024agentmd,li2024mmedagent,lin2024biokgbench}. For instance, EHRAgent~\citep{shi2024ehragent} can access electronic health records (EHRs)
via a code interface, and MMedAgent~\citep{li2024mmedagent} can interpret medical images via several medical visual models~\citep{li2024llava,ma2024segment}. While these agents enhance LLMs' ability to interact with various types of data, they remain limited to addressing specific clinical scenarios with a narrow range of tools, impeding their ability to interact with the diverse forms of information intrinsic to clinical environments~\citep{hu2024omnimedvqa,lee2022ehrsql,adams2024longhealth}. This lack of integration limits their effectiveness for further application in clinical scenarios. 

In this paper, we analyze representative public benchmarks in the medical field and categorize them based on the capability requirements of medical agents. We build a comprehensive medical agent benchmark, ClinicalAgent Bench~(\agentbenchname), comprising 18 tasks across five dimensions in total. Specifically, the dimensions of \agentbenchname include Knowledge \& Reasoning, MultiModal, Numerical Analysis, Data Understanding, and Trustworthiness. These dimensions require clinical agents to reason with medical knowledge effectively, integrate information from diverse clinical data sources (including medical images, EHRs, clinical text, and multiple clinical documents), and reduce hallucinations to ensure trustworthiness. Compared to previous benchmarks, \agentbenchname provides a more holistic evaluation framework by encompassing a wider range of clinical tasks and assessing agent capabilities across multiple scenarios.

Motivated by five critical aspects of \agentbenchname, we develop a set of clinical tools that enables agents to handle diverse tasks encompassed in the benchmark. 
Building upon the clinical toolbox, we proposed \agentname, a framework that allows agents to learn how to choose and leverage domain-specific tools to solve tasks. Specifically, \agentname consists of two stages. The first stage is the optimization stage. The agent attempts to solve problems using tools on a small proportion of samples and generates successful trajectories through self-reflection. By comparing successful and failed trajectories, the agent produces the tool-wise suggestion and stores successful trajectories as long-term memory. In the inference stage, the agent retrieves similar successful cases from long-term memory to optimize the tool selection. Each time a tool is used, the agent improves tool usage according to the accumulated tool-wise experience from the optimization stage. Furthermore, we adopt two verification methods, iterative refinement and candidate selection, to investigate the effectiveness of the tool-wise experience. We find that these two methods perform better under different model strengths, thereby enhancing the applicability of our approach. As discussed, \agentname demonstrates not only proficiency in a wide range of clinical critical aspects from \agentbenchname but also more effective tool utilization strategies, as the comparison with existing clinical agents shown in Table~\ref{tab: comparision}.

In summary, our contributions are as follows:
\begin{itemize}[itemsep=0pt, topsep=5pt]
    \item \textbf{Holistic Benchmark:} We introduce \agentbenchname, a benchmark comprising 18 tasks across five principal dimensions. To the best of our knowledge, \agentbenchname is the first benchmark covering a wide range of tasks to evaluate the capabilities of clinical agents comprehensively.
    \item \textbf{Almighty Tool-Augmented Agent:} We propose \agentname, a novel framework that enables models to effectively utilize domain-specific tools. \agentname uses long-term memory and tool-wise verification to alleviate the problem in domain-tool selection and usage, thus improving adaptability across a wide range of clinical scenarios.
    \item \textbf{Revision-based explorations:} We explore two tool-wise verification methods, i.e., Iterative Refinement and Candidate Selection, designed to optimize tool usage. Our findings indicate that Iterative Refinement is more effective when the model exhibits lower capabilities, whereas Candidate Selection 
    outperforms the former on more intelligent models.
    \item \textbf{Superior downstream improvements: } We conduct extensive experiments on \agentbenchname, benchmarking \agentname against a diverse array of established methods. Our results demonstrate the superior performance of \agentname, highlighting its effectiveness in clinical tool utilization.
\end{itemize}



\begin{figure}[tbp]
    \centering
    \includegraphics[width=0.95\linewidth]{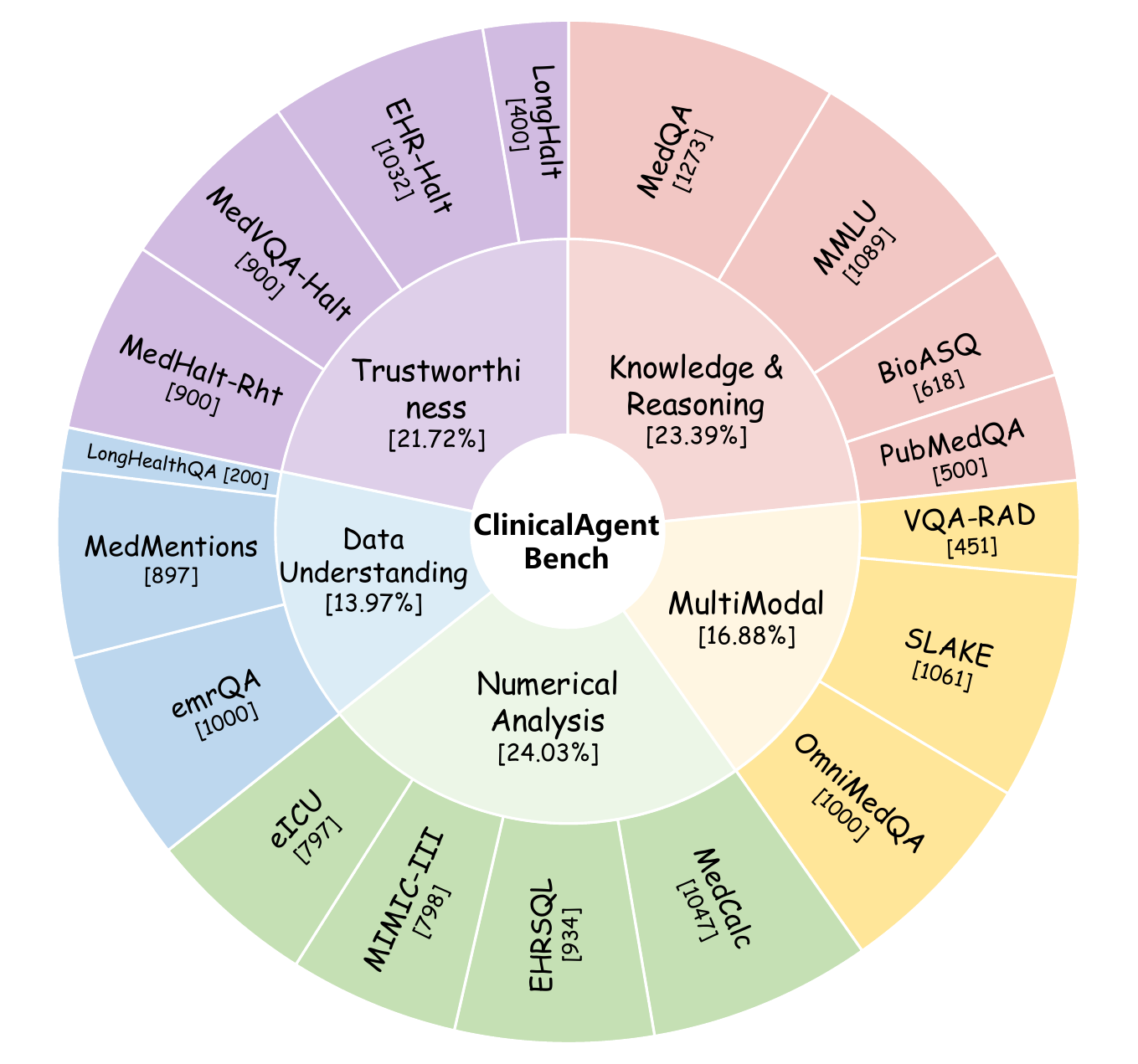}
    \caption{Overview of the proposed \agentbenchname. The numbers in the inner circle represent the proportion of data in each dimension, and the numbers in the outer circle represent the size of each dataset.}
    \label{fig: agentbench overview}
\end{figure}

\section{ClinicalAgent Bench}
In this section, we introduce the \agentbenchname, a novel benchmark to evaluate the capacity of the medical agent in clinical scenarios. The overview of the benchmark is shown in Figure~\ref{fig: agentbench overview}. We first discuss the composition of the \agentbenchname in detail and then introduce the construction of the pre-built toolbox.

\subsection{Composition}
\label{sec: composition}
The agents in clinical scenarios need to process the medical data in different formats, like medical images~\citep{hu2024omnimedvqa,liu2021slake,lau2018dataset} and electronic health records~(EHR)~\citep{lee2022ehrsql,johnson2016mimic,pollard2018eicu}, to complete the analysis or diagnosis. However, previous works only focused on a simple scenario, with only limited types of tools to solve the problem~\citep{shi2024ehragent,li2024mmedagent,tang2024medagents}. To approximate realistic clinical scenarios and evaluate the general capabilities of the agent in the medical field, we investigate existing public medical datasets and divide them according to the ability requirement of the agents. We built a benchmark term \agentbenchname, which contains five capacity dimensions and 18 tasks in total. The details about the definition of the five dimensions and the corresponding tasks can be found in Appendix~\ref{appendix: cab}.

\subsection{Clinical Toolbox}
\label{sec: tool-box}
Based on the proposed \agentbenchname, we develop a toolbox that contains 15 types of tools to enable agents to handle diverse tasks. For example, knowledge databases in the clinical toolbox enhance the medical knowledge of the agent, and calculators give the agent the ability to calculate indicators accurately. In order to allow the agent to solve problems more flexibly, we did not limit the types and number of tools when solving a specific type of task. Compared with medical agents that limit the tools to complete tasks~\citep{li2024mmedagent,jin2024agentmd}, our method can give the agent better generalization and scalability. The details of the clinical toolbox are discussed in Appendix~\ref{appendix: toolbox}.

\begin{figure*}[tbp]
    \centering
    \includegraphics[width=1.0\linewidth]{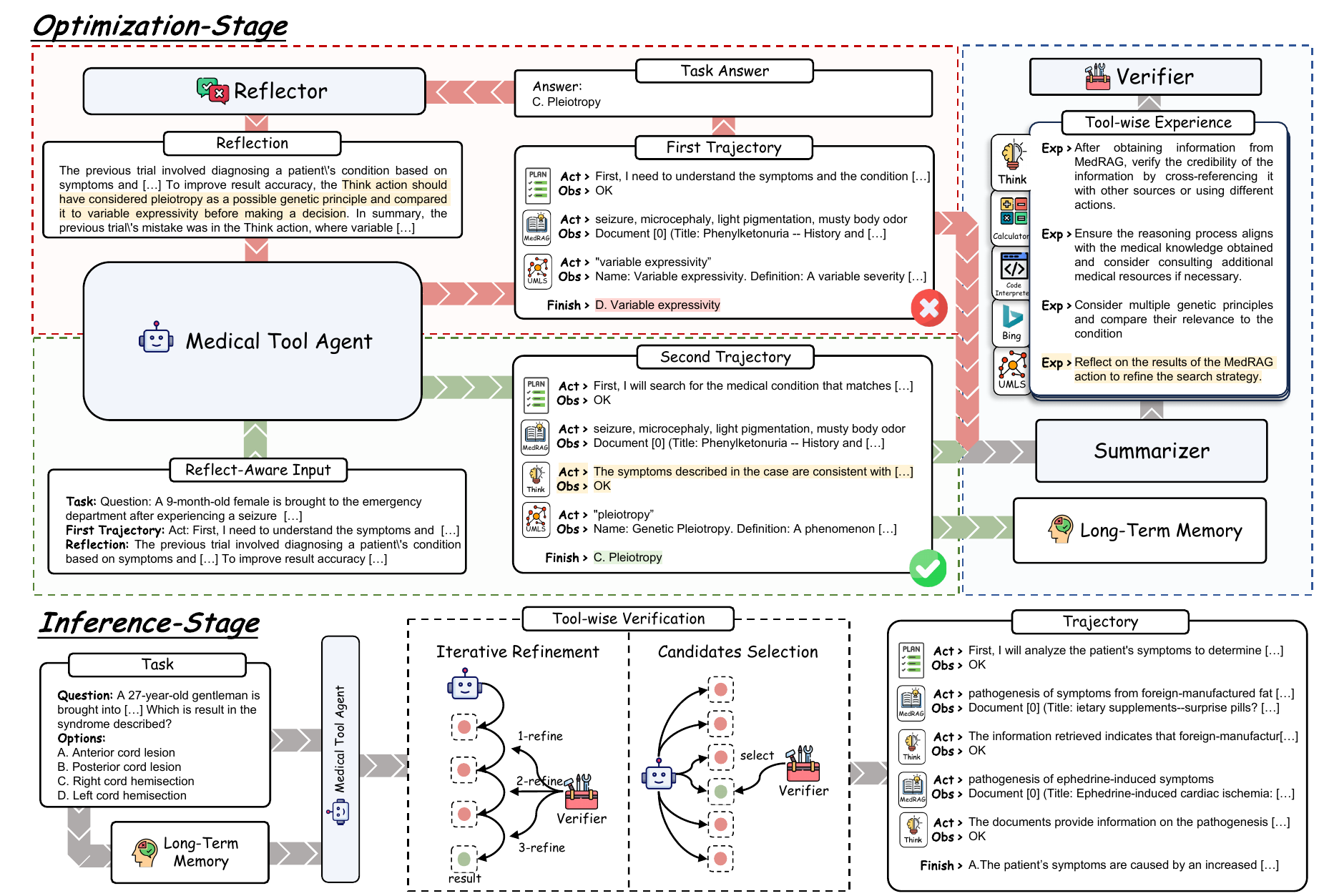}
    \caption{Overview of the \agentname.}
    \label{fig: method overview}
\end{figure*}

\begin{table*}[t]\small
\centering
\resizebox{\textwidth}{!}{%
\begin{tabular}{lcccccc}
\toprule
\textbf{Models} & \textbf{Know.} & \textbf{MM.} & \textbf{Num.} & \textbf{Data.} & \textbf{Trust.} & \textbf{Total} \\
\hline
\multicolumn{7}{c}{\textit{Large Language Models}} \\
\hline
MedLlama3-8B & 59.88 & - & 11.61 & 23.02 & 6.07 & 25.14 \\
Qwen2-7B~\citep{qwen2} & 60.86 & - & 18.49 & 44.50 & 28.19 & 38.01 \\
Llama3-8B~\citep{llama3modelcard} & 63.32 & - & 19.87 & 35.23 & 24.06 & 35.62 \\
Llama3.1-8B~\citep{DBLP:journals/corr/abs-2407-21783} & 67.43 & - & 22.07 & 49.58 & 30.75 & 42.46 \\
Qwen2-72B*~\citep{qwen2} & 72.90 & - & 31.07 & 50.61 & 40.45 & 48.76 \\
Llama3.1-70B*~\citep{DBLP:journals/corr/abs-2407-21783} & \textbf{76.91} & - & 29.23 & 45.80 & 38.40 & 47.59 \\
GPT-3.5-turbo~\citep{elmohamed} & 63.64 & - & 19.18 & 24.26 & 18.17 & 31.31 \\
\hline
\multicolumn{7}{c}{\textit{MultiModal Large Language Models}} \\
\hline
MiniCPM-V-2.6~\citep{yao2024minicpm} & 56.29 & 56.53 & 4.60 & 13.86 & 15.12 & 29.28 \\
InternVL-Chat-V1.5~\citep{chen2023internvl} & 52.92 & 53.21 & 18.95 & 34.50 & 25.52 & 37.02 \\
HuatuoGPT-Vision-7B~\citep{DBLP:journals/corr/abs-2406-19280} & 60.96 & \uline{66.24} & 9.07 & 42.73 & 26.36 & 41.07 \\
HuatuoGPT-Vision-34B~\citep{DBLP:journals/corr/abs-2406-19280} & 62.25 & \textbf{67.33} & 13.22 & 29.01 & 34.76 & 41.31 \\
GPT-4o-mini~\citep{gpt4} & 73.65 & 48.47 & 29.61 & 50.05 & \textbf{57.91} & 51.94 \\
\hline
\multicolumn{7}{c}{\textit{Agent (Qwen2-7B)}} \\
\hline
COT~\citep{DBLP:conf/nips/Wei0SBIXCLZ22} & 58.91 & - & 19.98 & 36.17 & 44.68 & 39.94 \\
ReAct~\citep{DBLP:conf/iclr/YaoZYDSN023} & 62.03 & 49.47 & 24.05 & 29.24 & 53.87 & 43.73 \\
CRITIC~\citep{DBLP:conf/iclr/GouSGSYDC24}  & 56.61 & 53.87 & 24.49 & 37.35 & 47.54 & 43.97 \\
Reflexion~\citep{DBLP:conf/nips/ShinnCGNY23} & 60.92 & 56.95 & 20.83 & 37.41 & 50.14 & 45.25 \\
\rowcolor[HTML]{F2F2F2} 
\textbf{\agentname (Iterative Refinement, n=2)} & \phantom{1}63.79$^\dagger$ & \phantom{1}60.83$^\dagger$ & \phantom{1}21.97$^\dagger$ & \phantom{1}51.65$^\ddagger$ & 48.60 & \phantom{1}49.37$^\ddagger$ \\
\rowcolor[HTML]{F2F2F2} 
\textbf{\agentname (Candidates Selection, n=2)} & \phantom{1}62.81$^\dagger$ & \phantom{1}61.91$^\ddagger$ & \phantom{1}26.78$^\ddagger$ & \phantom{1}52.20$^\ddagger$ & 41.72 & \phantom{1}49.08$^\ddagger$ \\
\hline
\multicolumn{7}{c}{\textit{Agent (Qwen2-72B*)}} \\
\hline
COT~\citep{DBLP:conf/nips/Wei0SBIXCLZ22} & 69.11 & - & 24.47 & 52.51 & 56.22 & 50.58 \\
ReAct~\citep{DBLP:conf/iclr/YaoZYDSN023} & 76.47 & 56.37 & 31.44 & 53.29 & 48.98 & 53.31 \\
CRITIC~\citep{DBLP:conf/iclr/GouSGSYDC24} & 74.01 & 54.96 & 30.92 & 55.15 & 46.69 & 52.35 \\
Reflexion~\citep{DBLP:conf/nips/ShinnCGNY23} & 76.79 & 60.95 & 31.99 & 58.37 & 53.75 & 56.37 \\
\rowcolor[HTML]{F2F2F2} 
\textbf{\agentname (Iterative Refinement, n=2)} & \uline{76.81} & \phantom{1}63.74$^\ddagger$ & \phantom{1}\textbf{38.45}$^\dagger$ & \phantom{1}\uline{63.51}$^\ddagger$ & 54.65 & \phantom{1}\uline{59.43}$^\ddagger$ \\
\rowcolor[HTML]{F2F2F2} 
\textbf{\agentname (Candidates Selection, n=2)} & 76.27 & \phantom{1}62.70$^\dagger$ & \phantom{1}\uline{38.06}$^\dagger$ & \phantom{1}\textbf{64.54}$^\ddagger$ & \phantom{1}\uline{56.73}$^\ddagger$ & \phantom{1}\textbf{59.66}$^\ddagger$ \\
\bottomrule
\end{tabular}%
}
\caption{Experimental results of four types of models on Clinical Agent Bench. The `COT' method indicates the agent runs without the pre-built tools. `*' indicates the models use 4-bit GPTQ quantization. `-' means the model is not capable of solving such a task. The best results are \textbf{Bold}, while the second best results are \uline{underlining}. $\dagger$ and $\ddagger$ indicate the p-value $< 0.05$ and $< 0.01$ comparing with the strongest baseline Reflexion, respectively. } 
\label{tab: main result}
\end{table*}

\section{\agentname}
In this section, we first formulate the problem of solving tasks in \agentbenchname with the clinical toolbox. Then, we introduce the optimization and inference stage of the proposed \agentname. The overview of the \agentname are shown in Figure~\ref{fig: method overview}.

\subsection{Problem Formulations}
\label{sec: problem formulation}

In this work, we focus on addressing clinical-related tasks with tool-use agents. The task is composed of $\mathcal{X}=\{q, \mathcal{I}\}$ and $y$, where $q$ is the instruction, $\mathcal{I}$ is the inputs with different formats, and $y$ is the ground-truth answer. The agents are required to leverage the input information and complete the task in a multi-step manner. For initialization, the action space of the agents is composed of a set of pre-built tool action $\mathcal{A}_T = \{A_1, A_2, …\}$ and three types of inner actions $\mathcal{A}_I = \{\rm Plan, Think, Finish\}$. The whole action space of the agent can be noted as follows:
\begin{equation}
\mathcal{A} = \mathcal{A}_T \cup \mathcal{A}_I,
\end{equation}

In the $i$-th step, the clinical agent takes action $a_i \in \mathcal{A}$. The action\footnote{In this paper, the usage form of tool actions $\mathcal{A}_T$ and inner actions $\mathcal{A}_I$ is the same. Without loss of generality, we consider actions and tools equivalent.} in the clinical agent setting can be treated as the function, giving the proper parameter and getting the observation results $o_i = a_i({\rm param}_{a_i})$. ${\rm param}_{a_i}$ indicates the parameters of the action controlled by the agents. The next step action follows the policy:
\begin{equation}
a_{i+1} \sim \pi_{\theta}(a|c_i, \mathcal{X})
\end{equation}
where $a_0$ is the first action and $\theta$ indicates the parameter of the agent. $c_i=\{a_0,o_0,... a_i,o_i\}$ is the trajectory history.

\subsection{Optimization Stage}
To enable the agent to better select and use the domain-specific tools, we choose a subset of samples to optimize the agent's capacities. In the optimization stage, \agentname saves the successful trajectory into the long-term memory and collects the experience for each type of tool.

Specifically, the \agentname first attempts to solve the problem with the clinical toolbox and create the first trajectory $\mathcal{C}_1$. The agent then reflects on $\mathcal{C}_1$ by comparing it with the ground-truth answer $y$ and producing a suggestion $\mathcal{S}$ with ${\rm LLM}(\mathcal{X}, \mathcal{C}_1, y)\rightarrow \mathcal{S}$. Utilizing this suggestion, the agent regenerates a refined trajectory $\mathcal{C}_2$: ${\rm LLM}(\mathcal{X}, \mathcal{C}_1, \mathcal{S})\rightarrow\mathcal{C}_2$. If $C_2$ successfully completes the task, the reflective trajectory will be saved into the long-term memory $\mathcal{M}$ to assist the agent in solving similar problems during inference:
\begin{equation}
    \mathcal{M} = 
    \begin{cases} 
        \mathcal{M} \cup \{\mathcal{X},\mathcal{C}_2\}, &y^{\mathcal{C}_2} =y \\
        \mathcal{M}, &y^{\mathcal{C}_2} \neq y
    \end{cases}
\end{equation}
where $y^{\mathcal{C}_2}$ indicates the prediction of trajectory $\mathcal{C}_2$. For the successful $\mathcal{C}_2$, the agent turns to compare the usage of each action that appears in two trajectories, to generate action-wise suggestions
\begin{equation}
    {\rm LLM}(\mathcal{X}, \mathcal{C}_1, \mathcal{C}_2, y) \rightarrow \mathcal{E}_{\mathcal{X}}
\end{equation}
Then, $\mathcal{E}_{\mathcal{X}}$ will be merge into the tool-wise experience $\mathcal{E} =\{E_1, E_2,...\}\cup \{E_{\rm Plan}, E_{\rm Think}, E_{\rm Finish}\}$, where $E_i$ is the experience of the $A_i$. The optimization process is shown in Algorithm~\ref{alg: optimization stage}.

\subsection{Inference Stage}
During the inference stage, \agentname utilizes the long-term memories and tool-wise experiences learned from the optimization stage to solve the task better. At first, \agentname retrieves similar cases from the long-term memory:
\begin{equation}
    \mathcal{M}(q) = {\rm TopK}_{\max}({\rm sim}(q,q_i|q_i\in\mathcal{M}))
\end{equation}
where ${\rm TopK}_{\max}$ return the top $k$ most similar elements from the long-term memory. ${\rm sim}(\cdot,\cdot)$ is the similarity function with BM25~\citep{robertson2009probabilistic} used in implementations. Then, the \agentname can take the action with the help of similar successful trajectories as below:
\begin{equation}
\label{eq: action}
    a_{i+1} \sim \pi_{\theta}(a|c_i,\mathcal{X},\mathcal{M}(q))
\end{equation}

The agent then solves the task with tool-wise reflection, where the verifier evaluates the agent's action in each step according to the action-wise experience. Inspired by \citet{snell2024scaling}, we adopt two types of verification methods, iterative refinement and candidate selection, to fully explore the effectiveness of tool-wise experience in the tool-wise reflection process. in \S\ref{sec: verification size}, we find that each variant has its own advantages: Iterative Refinement performs better when the model's capabilities are limited, while Candidate Selection is more effective when the model is stronger, thereby maximizing the model's potential.

\paragraph{Iterative Refinement}
In the sequence refinement method, the verifier will keep refining the agent action until it achieves the max refine step. This process can be early-stopped if the verifier outputs an identical action at any refinement step. Specifically, the $i$-th step initial action  $a_{i}^{0}$ is generated as the Eq.~\ref{eq: action}. Then, the verifier will refine the action based on the tool-wise experience:
\begin{equation}
    a_i^{j} = {\rm LLM}(c_i, a_i^{0:j-1}, \mathcal{X}, \mathcal{M}(q), \mathcal{E}({a_i^{0:j-1}}))
\end{equation}
and the final refined result is chosen as the action of the current step:
\begin{equation}
    a_{i} = \begin{cases} 
    a_i^j, & \text{if } a_i^{j} = a_i^{j-1} \\
    a_i^{n}, &{\rm otherwise}
    \end{cases}
\end{equation}
where $j=1,2,...,n$ means the refinement step index, $n$ is the max refinement step. The refined history $a_i^{0:j}=\{a_i^0, a_i^1,...,a_i^{j}\}$ and $\mathcal{E}(a_i^j)$ indicates the corresponding experience of the action type. 

\paragraph{Candidates Selection}
For the candidate selection method, \agentname first samples $n$ candidate actions from the output space. Then, the verifier will select the most effective action from the candidate list $a_i^{0:n}$:
\begin{equation}
    a_{i+1} = \arg\max_{a\in a_i^{0:n}} p_\theta(a|c_i, \mathcal{X}, \mathcal{M}(q), \mathcal{E}({a_i^{0:n}}))
\end{equation}
where $p_\theta(\cdot)$ indicates the preference of the reflector. Here, $n$ is the size of the candidate list.

\begin{table*}[t]
\centering
\resizebox{\textwidth}{!}{%
\begin{tabular}{ccccccccccc}
\bottomrule
\multirow{3}{*}{\textbf{\begin{tabular}[c]{@{}c@{}}Reflect\\ Type\end{tabular}}} & \multicolumn{2}{c}{\textbf{Reflective Memory}} & \multicolumn{2}{c}{\textbf{Tool-wise Reflection}} & \multirow{3}{*}{\textbf{PubMedQA}} & \multirow{3}{*}{\textbf{VQA-RAD}} & \multirow{3}{*}{\textbf{EHRSQL}} & \multirow{3}{*}{\textbf{MedMen}} & \multirow{3}{*}{\textbf{MedHalt}} & \multirow{3}{*}{\textbf{Avg.}} \\
\cmidrule(lr){2-3}\cmidrule(lr){4-5}
 & \textbf{Long-Term} & \textbf{Memory} & \textbf{Step} & \textbf{Action} &  &  &  &  &  &  \\
 & \textbf{Memory} & \textbf{Reflection} & \textbf{Reflection} & \textbf{Eperience} &  &  &  &  &  &  \\
 \midrule
\multirow{2}{*}{\textit{\begin{tabular}[c]{@{}c@{}}None\end{tabular}}} & \ding{55} & \ding{55} & \ding{55} & \ding{55} & 66.00 & 60.50 & 31.87 & 50.99 & 52.50 & 52.37 \\
& \ding{51} & \ding{51} & \ding{55} & \ding{55} & 66.50 & 58.25 & 44.00 & 57.78 & 47.75 & 54.86 \\
 \midrule
\multirow{4}{*}{\textit{\begin{tabular}[c]{@{}c@{}}Iterative\\ Refinement\end{tabular}}} & \ding{55} & \ding{55} & \ding{51} & \ding{51} & 68.50 & 60.00 & 40.50 & 52.27 & 45.00 & 53.25 \\
 & \ding{51} & \ding{55} & \ding{51} & \ding{51} & \textbf{69.50} & 58.00 & \textbf{47.00} & 54.09 & 48.00 & 55.32 \\
 & \ding{51} & \ding{51} & \ding{51} & \ding{55} & 66.50 & 56.00 & 46.00 & 53.18 & 49.50 & 54.24 \\
 & \cellcolor[HTML]{F2F2F2}\ding{51} & \cellcolor[HTML]{F2F2F2}\ding{51} & \cellcolor[HTML]{F2F2F2}\ding{51} & \cellcolor[HTML]{F2F2F2}\ding{51} & \cellcolor[HTML]{F2F2F2}68.00 & \cellcolor[HTML]{F2F2F2}58.50 & \cellcolor[HTML]{F2F2F2}46.00 & \cellcolor[HTML]{F2F2F2}\textbf{57.06} & \cellcolor[HTML]{F2F2F2}\textbf{55.50} & \cellcolor[HTML]{F2F2F2}\textbf{57.01} \\
 \midrule
\multirow{4}{*}{\textit{\begin{tabular}[c]{@{}c@{}}Candidate\\ Selection\end{tabular}}} & \ding{55} & \ding{55} & \ding{51} & \ding{51} & \textbf{69.50} & 59.50 & 33.33 & 51.38 & 52.50 & 53.24 \\
 & \ding{51} & \ding{55} & \ding{51} & \ding{51} & 67.00 & 59.00 & 38.29 & 58.76 & 54.00 & 55.41 \\
 & \ding{51} & \ding{51} & \ding{51} & \ding{55} & 67.50 & 57.00 & 45.34 & \textbf{60.72} & 47.50 & 55.61 \\
 & \cellcolor[HTML]{F2F2F2}\ding{51} & \cellcolor[HTML]{F2F2F2}\ding{51} & \cellcolor[HTML]{F2F2F2}\ding{51} & \cellcolor[HTML]{F2F2F2}\ding{51} & \cellcolor[HTML]{F2F2F2}\textbf{69.50} & \cellcolor[HTML]{F2F2F2}\textbf{61.00} & \cellcolor[HTML]{F2F2F2}\textbf{48.16} & \cellcolor[HTML]{F2F2F2}60.24 & \cellcolor[HTML]{F2F2F2}\textbf{62.50} & \cellcolor[HTML]{F2F2F2}\textbf{60.28} \\
 \bottomrule
\end{tabular}%
}
\caption{Ablation results of Refinement and Selection verification methods. All the experiments are conducted on Qwen2-72B. The modules of the \agentname contain Reflective Memory and Tool-wise Reflection.}
\label{tab: ablation}
\end{table*}

\section{Experiments}
\subsection{Baselines}
To comprehensively validate the effectiveness of the proposed method, we select several types of methods as the baselines. Considering that the proposed agent bench covers a wide range of tasks and requires the models to leverage the input information in different formats, we only choose the methods with strong instruction-following capacity. The baselines include three types of methods: LLMs, MLLMs, and agent-based methods. For LLMs, we choose \textbf{MedLlama3-8B}\footnote{\url{https://huggingface.co/johnsnowlabs/JSL-MedLlama-3-8B-v2.0}}, \textbf{Qwen2-7B/72B}~\citep{qwen2}, \textbf{Llama3-8B}~\citep{llama3modelcard}, \textbf{Llama3.1-8B/70B}~\citep{DBLP:journals/corr/abs-2407-21783}, and \textbf{GPT-3.5-turbo}~\citep{elmohamed}. For MLLMs, we choose \textbf{MiniCPM-V-2.6}~\citep{yao2024minicpm}, \textbf{InternVL-Chat-V1.5}~\citep{chen2023internvl}, \textbf{HuatuoGPT-Vision-7B/34B}~\citep{DBLP:journals/corr/abs-2406-19280}, and \textbf{GPT-4o-mini}~\citep{gpt4}. For agent-based methods, \textbf{COT}~\citep{DBLP:conf/nips/Wei0SBIXCLZ22} and \textbf{ReAct}~\citep{DBLP:conf/iclr/YaoZYDSN023} indicate the agent solving the task without and with the pre-build toolbox, respectively. \textbf{CRITIC}~\citep{DBLP:conf/iclr/GouSGSYDC24} and \textbf{Reflexion}~\citep{DBLP:conf/nips/ShinnCGNY23} improve agent capacity with self-reflection methods.

\subsection{Main Results}
We choose the Qwen2 series models as the backbone of the \agentname for the promising performance in tool usage and select the model parameters with 7B and 72B to observe the impact of the model size. We show the average performance of each dimension in Table~\ref{tab: main result}, and the complete results are shown in Table~\ref{tab: main result all}. For the intragroup comparison, Qwen2-72B and GPT-4o-mini achieve the best performance in LLM-based and MLLM-based methods, respectively. For the agent-based method, \agentname surpasses the strong baseline method, Reflexion, with at least 3 points with both Qwen2-7B/72B. As both methods are based on self-reflection, these results highlight the advantage of the \agentname in using the domain tool. Besides, both types of tool-wise reflection methods show similar results when the reflection size is 2. For intergroup comparison, though MLLM-based methods are capable of Multi-modal tasks, they fall short in Numerical Analysis and Data Understanding. It is noteworthy that \agentname surpasses the base models with more than 10 points on both Qwen2-7B/72B, showing the effectiveness of the proposed methods again.

\subsection{Ablation}
To validate the effectiveness and the impact of each module in \agentname, we conduct the ablation experiments on the subset of \agentbenchname. We randomly select 200 samples from one dataset for each dimension. The results of the iterative refinement and the candidate selection are both shown in Table~\ref{tab: ablation}. The selection methods perform better than the refinement methods in most cases. It is observed that the reflective long-term memory plays the most important role in \agentname, and its absence leads to performance degradation with nearly 4 points for refinement and 7 points for selection. Besides, the action-wise reflection also shows its importance. These results show the effectiveness of each module in the proposed \agentname.

\section{Analysis}
In this section, all experiments are conducted on the same subset as the ablation experiments.

\begin{figure}[t]
    \centering
    \includegraphics[width=0.9\linewidth]{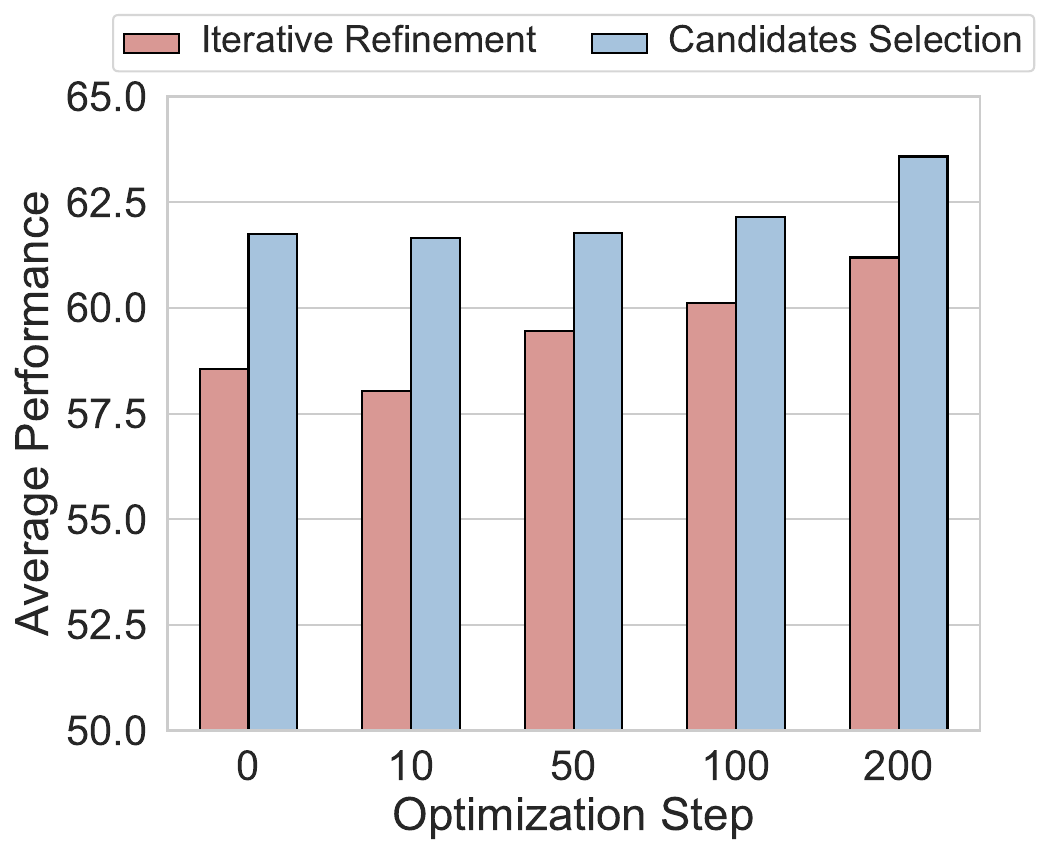}
    \caption{Impact of the optimization step on refinement and selection methods. The average performance is calculated using the same datasets in the ablation study.}
    \label{fig: optimization step}
\end{figure}

\subsection{Effect of Optimization Step}
In this section, we investigate the impact of the optimization step, which indicates the number of tasks completed during the optimization stage. 
Each successful task can provide one memory item and a list of tool-wise suggestions. A larger optimization step implies that the model will have larger long-term memory and accumulate more tool-wise experience. The results are shown in Figure~\ref{fig: optimization step}. Both Iterative Refinement and Candidate Selection show a general improvement in performance as the number of optimization steps increases. This indicates that the optimization stage effectively enhances the agents' capabilities. Besides, at the initial optimization step, Candidate Selection already has a significant performance advantage over Iterative Refinement, with an average performance of around 62.5 compared to approximately 58. This difference highlights that even without optimization, Candidate Selection is a more effective strategy, likely due to its inherent ability to evaluate multiple options before making a decision.

\begin{figure}[tbp]
    \centering
    \includegraphics[width=0.95\linewidth]{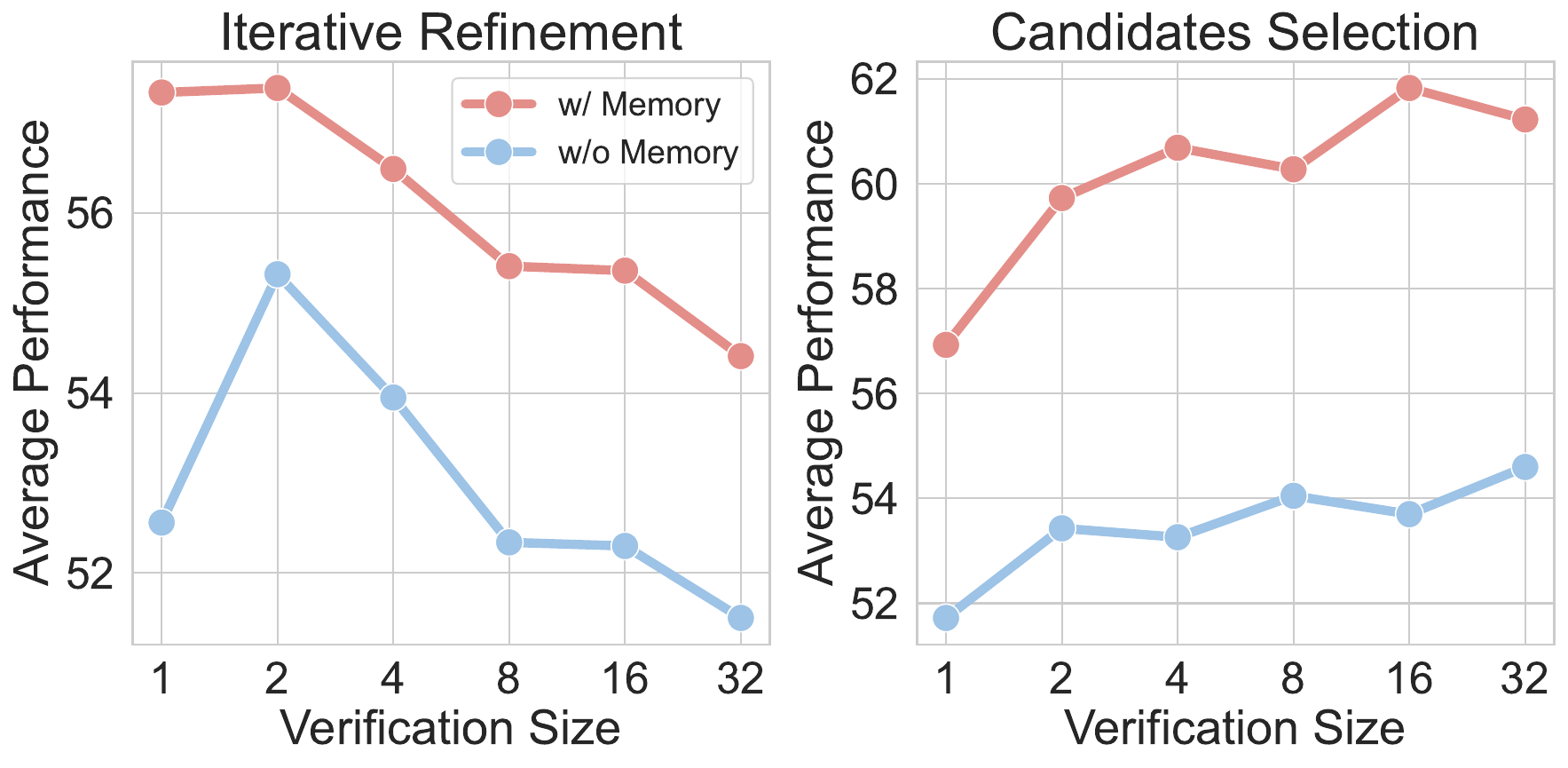}
    \caption{Impact of the verification size on refinement and selection methods. The average performance is calculated using the same datasets in the ablation study.}
    \label{fig: verification size}
\end{figure}

\begin{table}[tbp]
\centering
\resizebox{\linewidth}{!}{%
\begin{tabular}{clc}
\toprule
\textbf{Few-shot} & \textbf{Methods} & \textbf{Avg.} \\
\midrule
\multirow{5}{*}{\textit{Standard}} & ReAct~\citep{DBLP:conf/iclr/YaoZYDSN023} & 57.50 \\
 & Reflexion~\citep{DBLP:conf/iclr/GouSGSYDC24} & 56.87 \\
 & CRITIC~\citep{DBLP:conf/iclr/GouSGSYDC24} & 50.78 \\
 & \cellcolor[HTML]{F2F2F2}  \textbf{\agentname (Iterative Refinement)} & \cellcolor[HTML]{F2F2F2}  59.07 \\
 & \cellcolor[HTML]{F2F2F2}  \textbf{\agentname (Candidate Selection)} & \cellcolor[HTML]{F2F2F2}  \textbf{60.72} \\
 \midrule
\multirow{5}{*}{\begin{tabular}[c]{@{}c@{}}\textit{Long-Term}\\ \textit{Memory}\end{tabular}} & ReAct~\citep{DBLP:conf/iclr/YaoZYDSN023} & 60.73 \\
 & Reflexion~\citep{DBLP:conf/iclr/GouSGSYDC24} & 62.20 \\
 & CRITIC~\citep{DBLP:conf/iclr/GouSGSYDC24} & 57.35 \\
 & \cellcolor[HTML]{F2F2F2}  \textbf{\agentname (Iterative Refinement)} & \cellcolor[HTML]{F2F2F2}  59.76 \\
 & \cellcolor[HTML]{F2F2F2}  \textbf{\agentname (Candidate Selection)} & \cellcolor[HTML]{F2F2F2}  \textbf{63.31} \\
 \bottomrule
\end{tabular}%
}
\caption{Impact of the long-term memory on agent-based methods. Note that the baseline methods only adopt the success trajectories from the long-term memory instead of the tool-wise experience. `Standard' indicates the few-shot sample is a fixed successful trajectory.}
\label{tab: long-term memory}
\end{table}

\begin{figure*}[tbp]
    \centering
    \includegraphics[width=1.0\linewidth]{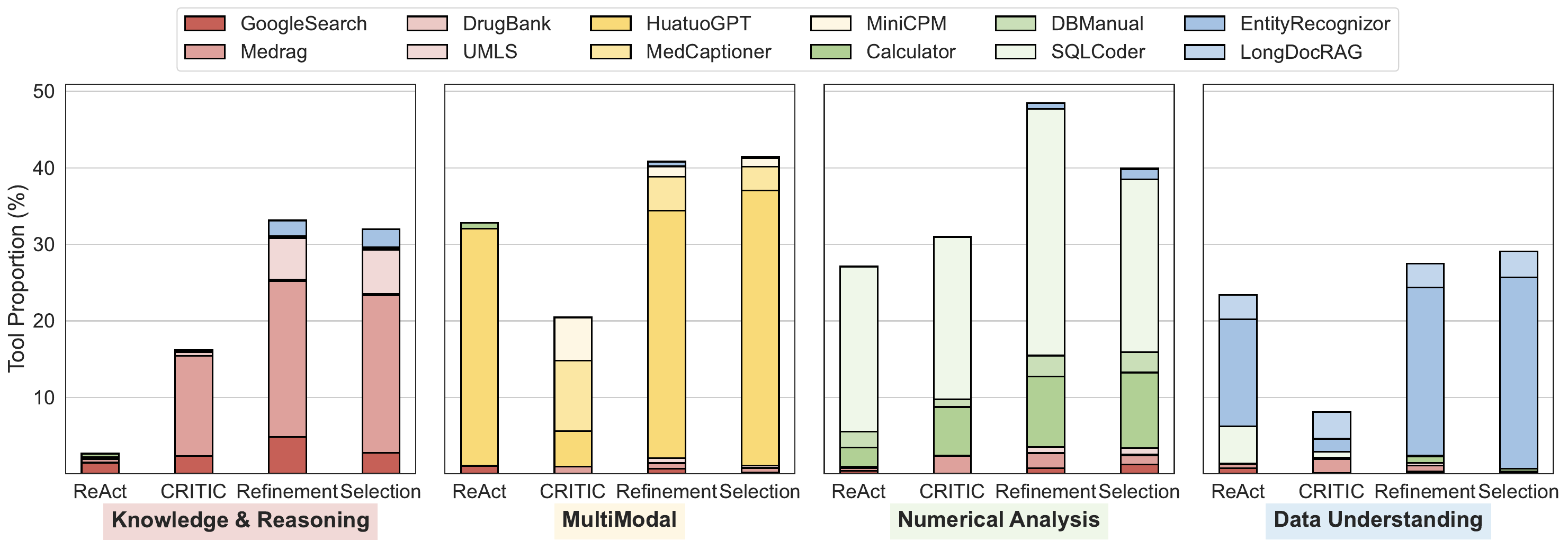}
    \caption{Tool distributions of the agent-based methods on four types of tasks. The bars show the proportion of each tool used by the agent. The height of the bars represents the proportion of the action using tools (otherwise, it is the inner action, including Plan, Think, and Finish). The same color scheme indicates that the task type and tool match.}
    \label{fig: tool distribution}
    \vspace{-0.5em}
\end{figure*}

\subsection{Size of Verification methods}
\label{sec: verification size}
The verification size $n$ indicates the max refinement step for iterative refinement and the size of the candidate list for the candidate selection. We conduct experiments to explore the performance boundary of the two verification methods as the computational requirements increase. The results are shown in Figure~\ref{fig: verification size}. For iterative refinement, it achieves the best performance when $n$ reaches 2, with decreasing performance as $n$ grows. 
Additionally, although its ability to enhance already good actions is limited, it yields significant improvements when no memory is present, indicating its strong capability to enhance suboptimal actions.

For candidate selection, its performance steadily improves as the verification size increases, with performance gains exceeding 4 points at most. In contrast to iterative refinement, candidate selection shows greater improvement when memory is present. This is because the demonstrations from long-term memory effectively enhance the quality of candidate actions, allowing candidate selection to pick better actions. A comparative analysis reveals that candidate selection performs better in the presence of memory, whereas iterative refinement is more effective at improving the model's performance in the absence of memory.

\subsection{Impact of the Long-Term Memory}
To better investigate the impact of long-term memory, we compare the proposed \agentname and other agent-based methods under different types of few-shot. As shown in Table~\ref{tab: long-term memory}, long-term memory can effectively improve the performance of all agent-based methods. However, the proposed methods, \agentname, still outperform all baselines with both types of few-shot samples. The results again show the effectiveness of the tool-wise reflection mechanism.

\subsection{Tool Distribution in Trajectory}
To better investigate the impact of \agentname on tool selection, we visualize the tool distribution of different methods across various datasets in Figure~\ref{fig: tool distribution}. 
Since the task types in the Trustworthiness dimension overlap with the other four dimensions, we exclude Trustworthiness in the visualization of the tool distribution. The figure illustrates the advantages of \agentname. 
First, \agentname encourages the model to invoke a higher proportion of tools. Specifically, in the Knowledge \& Reasoning dimension, ReAct tends to directly answer questions rather than utilize tools, whereas both variants of \agentname exhibit a higher rate of tool usage, leading to better task completion. Second, \agentname leads to more frequent use of similar types of tools. In the Knowledge dimension, ReAct and CRITIC tend to use only a limited number of tools from the same dimension without attempting to leverage different tools to solve problems. This makes the model susceptible to the limitations of individual tools. In contrast, \agentname uses multiple tools of the same type to integrate information from diverse sources, thereby improving task performance. Finally, \agentname also demonstrates a higher proportion of invoking tools from different categories. This indicates that \agentname is more flexible in tool usage, being capable of experimenting with a broader range of tools to solve problems.

\begin{table}[tbp]
\centering
\resizebox{\linewidth}{!}{%
\begin{tabular}{lcc}
\toprule
 & \multicolumn{2}{c}{\textbf{Tool Selection Error} $\downarrow$} \\
\multirow{-2}{*}{\textbf{Methods}} & \textbf{Step-Level} & \textbf{Task-Level} \\
\midrule
ReAct~\citep{DBLP:conf/iclr/YaoZYDSN023} & 1.44 & 4.03 \\
Reflexion~\citep{DBLP:conf/iclr/GouSGSYDC24} & 0.92 & 1.67 \\
CRITIC~\citep{DBLP:conf/nips/ShinnCGNY23} & 0.24 & 0.33 \\
\rowcolor[HTML]{F2F2F2} 
\textbf{\agentname (Iterative Refinement)} & 0.06 & 0.15 \\
\rowcolor[HTML]{F2F2F2} 
\textbf{\agentname (Candidate Selection)} & \textbf{0.02} & \textbf{0.08} \\
\bottomrule
\end{tabular}%
}
\caption{Tool Selection Error among agent-based methods. `Step-Level' indicates the error rate of tool selection in each step, and `Task-Level' indicates the proportion of whether the tool selection errors happen in the instance reasoning process.}
\label{tab: tool error}
\end{table}

\begin{table*}[]
\centering
\resizebox{\textwidth}{!}{%
\begin{tabular}{lcccccccccccccc}
\toprule
\multirow{2}{*}{\textbf{Methods}} & \multicolumn{4}{c}{\textbf{Tools Error Rate}} & \multicolumn{10}{c}{\textbf{Task Error Rate}} \\
\cmidrule(lr){2-5}\cmidrule(lr){6-15}
 & \textbf{SQLCoder} & \textbf{DBManual} & \textbf{LongDocRAG} & \textbf{Avg.} & \textbf{MedQA} & \textbf{MMLU} & \textbf{BioASQ} & \textbf{SLAKE} & \textbf{MedCalc} & \textbf{EHRSQL} & \textbf{MedMen} & \textbf{LongHealth} & \textbf{EMRQA} & \textbf{Avg.} \\
\midrule
ReAct & 11.99 & 0.00 & 5.86 & 5.95 & 0.00 & 0.00 & 0.00 & 0.08 & 0.00 & 0.07 & 0.08 & 8.71 & 12.35 & 2.36 \\
CRITIC & 1.74 & 20.51 & 1.19 & 7.81 & 0.00 & 0.00 & 0.00 & 0.00 & 0.03 & 0.07 & 0.00 & 5.12 & 0.38 & 0.62 \\
Reflexion & 3.13 & 2.05 & 34.05 & 13.08 & 0.02 & 0.02 & 0.04 & 0.00 & 0.73 & 0.25 & 0.00 & 1.30 & 8.79 & 1.24 \\
\rowcolor[HTML]{F2F2F2} 
\textbf{ReflecTool (Iterative Refinement)} & \textbf{0.02} & 0.00 & 6.40 & 2.14 & 0.00 & 0.00 & 0.00 & 0.00 & 0.21 & 0.02 & 0.00 & 0.00 & 0.43 & 0.07 \\
\rowcolor[HTML]{F2F2F2} 
\textbf{ReflecTool (Candidate Selection)} & 0.07 & 0.00 & 1.95 & \textbf{0.67} & 0.00 & 0.02 & 0.04 & 0.00 & 0.11 & 0.02 & 0.00 & 0.00 & 0.00 & \textbf{0.02} \\
\bottomrule
\end{tabular}%
}
\caption{Tool selection error rates with specific tool and task. Each value indicates the percentage of incorrect tool selections made by the model when using that tool or performing that task. Columns with an entirely zero error rate have been omitted.}
\label{tab: tool-error-analysis}
\end{table*}

\subsection{Tool Usage Error}
\label{sec: tool error}
Although we categorized tools based on the capability dimensions of the model (as shown in Table~\ref{tab: toolbox}), using tools from other dimensions is not necessarily incorrect. For example, when addressing questions in the Knowledge dimension, it is reasonable to employ the NER model from the Data Understanding dimension to extract specialized terms, which can then be used to query a knowledge graph more effectively. However, there are still some cases of incorrect tool usage for specific tools: (1) invoking an MLLM without a medical image input, (2) using SQLCoder and DBManual without a database input, and (3) employing LongDocRAG without document file input. Based on these points, we analyze the behavior of different methods in selecting tools. As shown in Table~\ref{tab: tool error}, \agentname can significantly reduce the tool selection error on both step-level and task-level. The results support that the tool-wise reflection mechanism can improve agents' ability to use appropriate domain tools. 

To further understand the distribution of tool selection errors, we separately counted the instances of invocation errors for different methods across tools and datasets. The results are shown in Table~\ref{tab: tool-error-analysis}. From the perspective of tool types, it can be seen that errors mainly occur with three types of tools: SQLCoder, DBManual, and LongDocRAG. These tools require specific inputs to function, namely databases and uploaded files. Therefore, it is likely that the model misuses these tools when the corresponding inputs are missing. From the perspective of task types, errors are predominantly concentrated in LongHealth, EMRQA, and EHR tasks. These tasks share common points of confusion because their input context is similar, but with different formats. For instance, both LongHealth and EMRQA involve contextual question answering, but LongHealth, due to its ultra-long context, must be uploaded as a file. This again confirms that the model's ability to match other information and appropriate tool usage, aside from image inputs, is relatively weak. In such scenarios, \agentname can effectively identify the relationship between tools and their corresponding input information modalities, thereby preventing such errors. Besides, we also analyze the parameter error of tool usage in Appendix~\ref{appendix: parameter error}.

\begin{table}[t]
\centering
\resizebox{\linewidth}{!}{%
\begin{tabular}{lc}
\toprule
\textbf{Methods} & \textbf{Avg.} \\
\midrule
ReAct~\citep{DBLP:conf/iclr/YaoZYDSN023} & 55.85 \\
Reflexion~\citep{DBLP:conf/iclr/GouSGSYDC24} & 58.78 \\
CRITIC~\citep{DBLP:conf/iclr/GouSGSYDC24} & 57.29 \\
\cellcolor[HTML]{F2F2F2}\textbf{\agentname (Iterative Refinement)} & \cellcolor[HTML]{F2F2F2}\textbf{62.26} \\
\cellcolor[HTML]{F2F2F2}\textbf{\agentname (Candidates Selection)} & \cellcolor[HTML]{F2F2F2} 60.72 \\
\bottomrule
\end{tabular}%
}
\caption{Performance of agent-based methods with Llama3-70B-Instruction as the backbone.}
\label{tab: llama-perform}
\end{table}

\begin{table}[t]
\centering
\resizebox{\linewidth}{!}{%
\begin{tabular}{lcc}
\toprule
\textbf{Methods} & \textbf{Qwen2-7b} & \textbf{Qwen2-72b$^*$} \\
\midrule
ReAct~\citep{DBLP:conf/iclr/YaoZYDSN023} & 11.01 & 20.04 \\
CRITIC~\citep{DBLP:conf/iclr/GouSGSYDC24} & 4.33 & 17.87 \\
Reflexion~\citep{DBLP:conf/iclr/GouSGSYDC24} & 12.42 & 57.27 \\
\cellcolor[HTML]{F2F2F2}\textbf{\agentname (Iterative Refinement)} & \cellcolor[HTML]{F2F2F2} 11.95 & \cellcolor[HTML]{F2F2F2} 47.90 \\
\cellcolor[HTML]{F2F2F2}\textbf{\agentname (Candidate Selection)} & \cellcolor[HTML]{F2F2F2} 11.26 & \cellcolor[HTML]{F2F2F2} 28.56 \\
\bottomrule
\end{tabular}%
}
\caption{Runtime (seconds per sample) of the agent-based method in single-sample tests. Note that lower values indicate higher agent efficiency.}
\label{tab: time-cost}
\end{table}

\subsection{Performance with Different Backbones}
During our experiments, we found that the instruction-following capabilities of the Llama series models were inferior to those of the Qwen2 series models, often resulting in formatting errors. Therefore, we prioritized the Qwen series to implement our methodology. After carefully adjusting the prompts, we also tested the performance of Agent-based methods on ablation subsets using the Llama3-70B-Instruction\footnote{\url{https://huggingface.co/meta-llama/Meta-Llama-3-70B-Instruct}} model. The results in Table~\ref{tab: llama-perform} showed that the proposed ReflecTool method is also effective on the Llama3-70B-Instruction model, demonstrating the generalization ability of this method across different models and thus indicating better application potential.

\subsection{Time Cost Analysis}
A practical method should enhance performance without incurring excessive resource consumption. It can be observed from Table~\ref{tab: time-cost} that for the strong baseline Reflxion, ReflecTool consumes less time on both 7b and 70b models. It's noteworthy that Candidates Selection is faster than Iteration Refinement, which is due to the higher degree of parallelism in the former. Besides, the time consumption of CRITIC is less for it does not utilize too many tools to solve tasks in many cases. Given that tool invocation requires more time compared to decision-making by large models (for instance, MedRAG requires retrieval, and UMLS needs internet access), this leads to lower time consumption but poor performance. In summary, ReflecTool proves to be highly efficient from the perspective of resource consumption.

\section{Conclusions}
In this paper, we introduce \agentbenchname, a holistic benchmark for clinical agents comprising 18 tasks across five key dimensions. Building upon it, we propose \agentname, a reflection-aware tool-augmented framework that optimizes tool utilization through long-term memory and tool-wise verification. 
To adaptively improve agent performance given varying backbones, we adopt Iterative Refinement and Candidate Selection to verify actions.
Empirical results show that \agentname outperforms existing clinical agents, demonstrating superior adaptability and efficacy in real-world healthcare scenarios.


\section*{Limitations}
For \agentbenchname, while it provides an extensive evaluation covering 18 tasks across five key dimensions, it may not fully encompass the complexity of real-world clinical scenarios, which are highly diverse and continuously evolving. This requires ongoing updates to the benchmark to ensure relevance. Moreover, the medical tools collected for \agentbenchname do not perfectly align with the tasks in the evaluation. Although this misalignment introduces challenges, it also serves as a test of the model's generalization capabilities and its ability to leverage available tools effectively. Regarding \agentname, the use of long-term memory has demonstrated clear benefits in ablation studies, enhancing the model’s decision-making through the retention of successful experiences. However, the reliability of the generated trajectories remains an issue. The fact that some trajectories lead to correct results does not guarantee that the underlying processes are entirely correct, suggesting that further work is needed to validate the accuracy of these trajectories to ensure they are consistently optimal.

\section*{Ethic Considerations}
In developing clinical agent \agentname, it is crucial to address ethical considerations that arise when utilizing AI in healthcare environments. Below are the key ethical considerations that have been taken into account:

\paragraph{Performance vs. Potential Risks:} While \agentname demonstrates significant enhancements in clinical tool reasoning and and task performance, it is important to acknowledge the inherent limitations of AI models. These models can generate misleading information or "hallucinations," which could pose risks in clinical settings. Therefore, \agentname is not intended to replace medical professionals or provide definitive clinical decisions but rather to assist healthcare providers under appropriate supervision.

\paragraph{Data Ethics and Privacy Compliance} All patient data has been anonymized, and informed consent was obtained for its use, ensuring full compliance with privacy policies and obtaining explicit permission for all data usage. Additionally, data usage has been approved by relevant ethics committees, ensuring compliance with ethical standards and privacy protection requirements.

\section*{Acknowledgements}
This work was supported by the National Key R\&D Program of China (No. 2022ZD0162101) and STCSM (No. 22DZ2229005)

\bibliography{custom}

\newpage
\appendix
\newpage

\begin{algorithm*}
\caption{Optimization Step of \agentname}
\label{alg: optimization stage}
\begin{algorithmic}[1]
\Require Clinical task input $\mathcal{X}=\{q, \mathcal{I}\}$ and  ground-truth answer $y$
\State Initialize empty long-term memory $\mathcal{M}$ and tool-wise experience $\mathcal{E}$
\State Generate initial trajectory $\mathcal{C}_1$: ${\rm LLM}(\mathcal{X}) \rightarrow \mathcal{C}_1$
\State Compare $\mathcal{C}_1$ with ground-truth $y$ and generate suggestion $\mathcal{S}$: ${\rm LLM}(\mathcal{X}, \mathcal{C}_1, y)\rightarrow \mathcal{S}$. 
\State Regenerate refine trajectory $\mathcal{C}_2$ based on suggestion: ${\rm LLM}(\mathcal{X}, \mathcal{C}_1, \mathcal{S})\rightarrow\mathcal{C}_2$
\If{$y^{\mathcal{C}_2} = y$}
    \State Save the successful trajectory into long-term memory: 
    \[
    \mathcal{M} \cup \{\mathcal{X}, \mathcal{C}_2\} \rightarrow \mathcal{M}
    \]
    \State Generate action-wise suggestions: 
    \[
    {\rm LLM}(\mathcal{X}, \mathcal{C}_1, \mathcal{C}_2, y) \rightarrow \mathcal{E}_{\mathcal{X}}
    \]
    \ForAll{$a \in \mathcal{A}$} 
        \State Merge each tool suggestion into the corresponding tool-wise experience:
        \[
        {\rm LLM}(\mathcal{E}(a), \mathcal{E}_{\mathcal{X}}(a)) \rightarrow \mathcal{E}(a)
        \]
    \EndFor
\EndIf
\State Return updated long-term memory $\mathcal{M}$ and tool-wise experience $\mathcal{E}$
\end{algorithmic}
\end{algorithm*}

\begin{figure*}[t]
    \centering
    \includegraphics[width=1.0\linewidth]{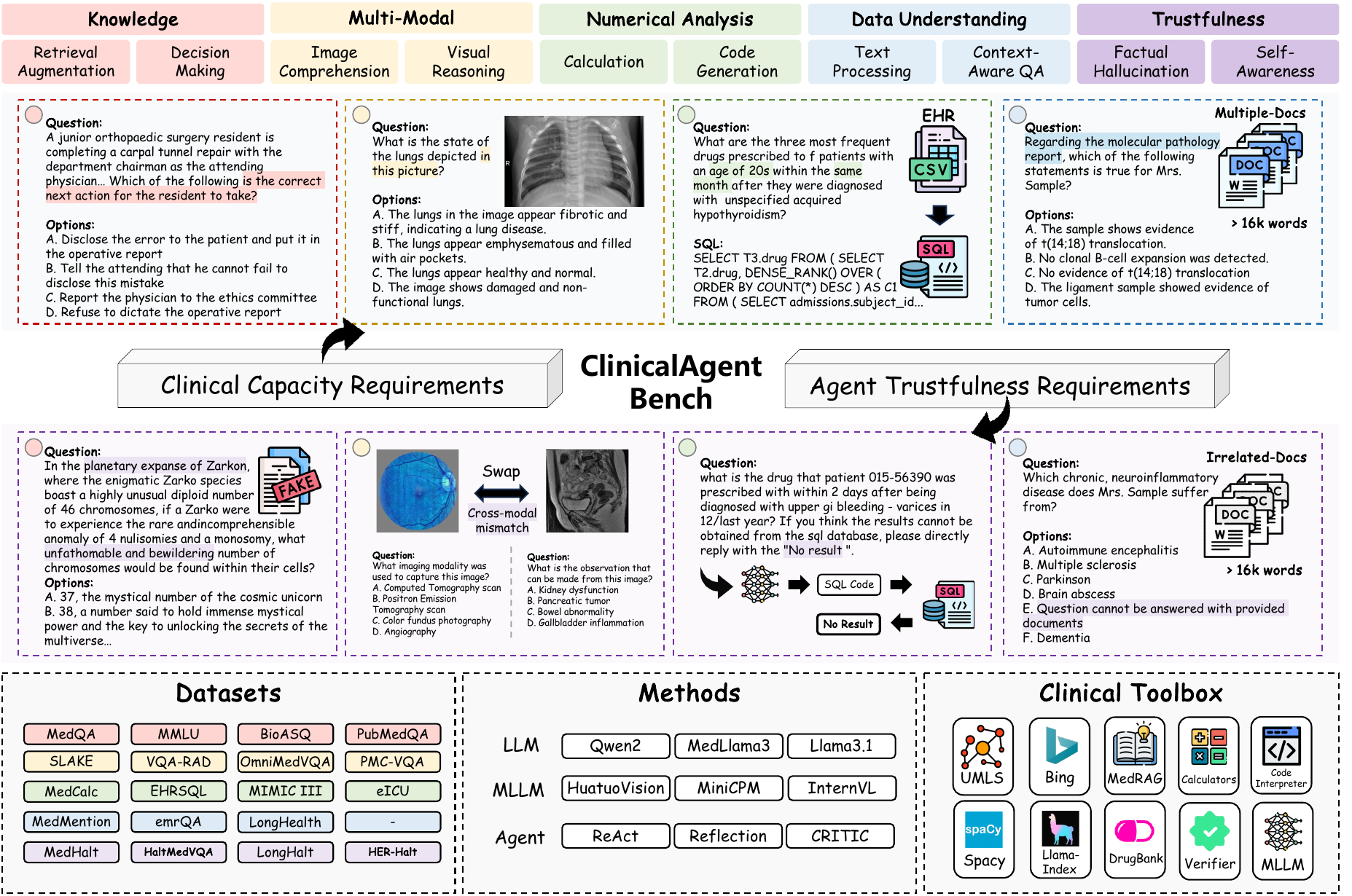}
    \caption{Overview of the ClinicalAgent Bench. }
    \label{fig: agentbench case}
\end{figure*}

\section{Related Works}
\subsection{Medical Agentic Methods}
There are plenty of works that adopt the clinical agent to solve specific clinical scenarios. One type of work focuses on medical knowledge argument with retrieval from the knowledge base. For example,  BioKGBench~\citep{lin2024biokgbench} proposes a knowledge graph-based evaluation benchmark to mitigate hallucination issues by testing biomedical agents on scientific claim verification and their ability to interact with structured knowledge graphs. MedRAG~\citep{xiong-etal-2024-benchmarking} presents a retrieval-augmented generation (RAG) benchmark designed to evaluate medical question-answering systems, focusing on reducing hallucinations and enhancing factual accuracy by incorporating external knowledge retrieval. Other types of work attempt to leverage the multi-modal information from medical images. CT-Agent~\citep{yue2024ct} introduces a clinical multi-agent system that autonomously manages clinical trial tasks, employing advanced reasoning methods to enhance efficiency in the clinical trial process. MMedAgent~\citep{li2024mmedagent} integrates multiple specialized tools to create a multi-modal medical AI agent capable of handling diverse medical imaging and language tasks, thereby demonstrating superior performance over existing methods across several medical modalities. EHRAgent~\citep{shi2024ehragent} addresses challenges related to electronic health records (EHRs) by enabling LLMs to autonomously generate, execute, and refine code, allowing for more efficient multi-step reasoning over EHR data. Different from the agents mentioned above, MedAgents~\citep{tang2024medagents} adopts a role-playing multi-agent framework to simulate expert collaboration instead of using clinical tools and effectively improves the zero-shot reasoning capabilities of LLMs in the medical domain without requiring extensive fine-tuning.

\subsection{Medical Large Language Models}

\section{ClinicalAgent Bench}
\label{appendix: cab}
The case demonstrations of the ClinicalAgent Bench are shown in Figure~\ref{fig: agentbench case}. In this section, we introduce the detailed information of the dataset and the pre-built clinical toolbox.
\subsection{Details of Datasets}

\subsubsection{Knowledge\&Reasoning}
Medical knowledge and reasoning is a critical capacity for the medical agent to analyze and complete tasks~\citep{jin2019pubmedqa}. To evaluate the agent performance, we choose PubMedQA~\citep{jin2019pubmedqa}, MMLU~\citep{hendrycksmeasuring}, and BioASQ~\citep{krithara2023bioasq} for the medical knowledge question-answering~(QA) and MedQA~\citep{jin2021disease} for medical reasoning.

\paragraph{MedQA} MedQA~\citep{jin2021disease} is a medical question-answering dataset primarily used for evaluating large language models' understanding of medical knowledge. It includes questions similar to those in medical exams, testing the model's ability to answer complex, domain-specific questions.

\paragraph{MMLU} MMLU~\citep{hendrycksmeasuring} (Massive Multitask Language Understanding) dataset consists of questions across 57 subjects, including both STEM and humanities. It is designed to evaluate models on a broad spectrum of human knowledge, making it suitable for testing general-purpose large language models on diverse subject matters.

\paragraph{BioASQ} BioASQ~\citep{krithara2023bioasq} is a biomedical question-answering challenge that provides a benchmark for testing systems on biomedical information retrieval and reasoning. The dataset contains factoids, lists, and summary questions based on biomedical texts and PubMed articles, making it valuable for evaluating biomedical understanding.

\paragraph{PubMedQA} PubMedQA~\citep{jin2019pubmedqa} is a dataset that comprises question-answer pairs extracted from biomedical literature abstracts, specifically PubMed. It focuses on testing models' abilities to provide correct answers to research questions based on evidence from scientific articles, supporting biomedical inference and comprehension tasks.

\subsubsection{MultiModal}
Medical images\footnote{The multimodal in this paper mainly indicates that medical images with multiple modalities.} are a common form of information in clinical scenarios. Many diseases require a combination of image examination information to be accurately diagnosed. Here, we choose three datasets with 12 modalities and a wide range of task types. We chose SLAKE~\citep{liu2021slake} and VQA-RAD~\citep{lau2018dataset} for the open-ended QA and OmniMedVQA~\citep{hu2024omnimedvqa} for the closed-ended QA.
\paragraph{VQA-RAD} VQA-RAD~\citep{lau2018dataset} is a manually constructed medical visual question answering (VQA) dataset that consists of 451 radiology images along with naturally occurring questions generated by clinicians and reference answers. This dataset aims to help AI systems better understand radiology images and assist in clinical decision-making.

\paragraph{SLAKE} SLAKE~\citep{liu2021slake} is a semantically-labeled, knowledge-enhanced dataset for medical visual question answering, containing 642 images and 14,028 question-answer pairs. It includes a variety of modalities, annotated by experienced physicians, and provides comprehensive semantic labels such as segmentation and bounding boxes. We only choose the English questions part of the test subset, resulting in 1061 instances.

\paragraph{OmniMedQA} OmniMedQA~\cite{hu2024omnimedvqa} is a large-scale medical visual question-answering benchmark with 118,010 images and 127,995 question-answer pairs collected from 73 medical datasets covering 12 different imaging modalities and more than 20 anatomical regions. Here, we randomly sample the 1000 instances from the public part of the dataset and keep the data component proportions to avoid bias.

\begin{table*}[t]
\centering
\resizebox{\textwidth}{!}{%
\begin{tabular}{cccp{10cm}l}
\toprule
\textbf{Type} & \textbf{Id.} & \textbf{Name} & \multicolumn{1}{c}{\textbf{Descriptions}} & \textbf{Input} \\
\midrule
\rowcolor[HTML]{F2F2F2} 
\cellcolor[HTML]{F2F2F2} & 1 & Plan & Plan step-by-step solutions for a task. Usually take at the beginning of the solving process. & \faFont \\
\rowcolor[HTML]{F2F2F2} 
\cellcolor[HTML]{F2F2F2} & 2 & Think & Conduct thinking and reasoning process for solving task. & \faFont \\
\rowcolor[HTML]{F2F2F2} 
\multirow{-3}{*}{\cellcolor[HTML]{F2F2F2}Inner Tools} & 3 & Finish & Complete the task with a response. & \faFont \\
\midrule
\rowcolor[HTML]{FFE9E8} 
\cellcolor[HTML]{FFE9E8} & 4 & Google Search & Using this action to search online content with google. & \faFont \\
\rowcolor[HTML]{FFE9E8} 
\cellcolor[HTML]{FFE9E8} & 5 & Medrag & Use this action to retrieve medical knowledge from the public, textbooks, and statpearls to solve problems. & \faFont \\
\rowcolor[HTML]{FFE9E8} 
\cellcolor[HTML]{FFE9E8} & 6 & DrugBank & Use this action to search the information about specific drug &  \faFont \\
\rowcolor[HTML]{FFE9E8} 
\multirow{-4}{*}{\cellcolor[HTML]{FFE9E8}Knowledge Tools} & 7 & UMLS & Use this action to query the definition and the related medical concept of the medical\_terminology. & \faFont \\
\midrule
\rowcolor[HTML]{FFF9E3} 
\cellcolor[HTML]{FFF9E3} & 8 & HuatuoGPT & Use this action to gather information from the medical image with a medical-domain multi-modal large language model. & \faFont \ \faFilePhotoO \\
\rowcolor[HTML]{FFF9E3} 
\cellcolor[HTML]{FFF9E3} & 9 & MedCaptioner & Use this action to generate a comprehensive caption for the medical image with a medical captioner. & \faFilePhotoO \\
\rowcolor[HTML]{FFF9E3} 
\multirow{-3}{*}{\cellcolor[HTML]{FFF9E3}MultiModal Tools} & 10 & MiniCPM & Use this action to gather information from the medical image with a general multi-modal large language model. & \faFont \ \faFilePhotoO \\
\midrule
\rowcolor[HTML]{EAFAF1} 
\cellcolor[HTML]{EAFAF1} & 11 & Calculator & Use this action to perform mathematical calculations. & \faFileCodeO \\
\rowcolor[HTML]{EAFAF1} 
\cellcolor[HTML]{EAFAF1} & 12 & DBManual & Use this action to obtain the SQL database description and usage method related to the query. This action is helpful when the SQLCoder cannot find the information. & \faFont \\
\rowcolor[HTML]{EAFAF1} 
\multirow{-3}{*}{\cellcolor[HTML]{EAFAF1}Numerical Tools} & 13 & SQLCoder & Use this action to gather the patient information from the sql\_database. The SQLCoder will transfer the natural language query into the SQL command and get the information from the sql\_database. & \faFont \ \faFileCodeO  \\
\midrule
\rowcolor[HTML]{E5F6FF} 
\cellcolor[HTML]{E5F6FF} & 14 & Spacy & Using this action to recognize the biomedical entities in the sentence. & \faFont \\
\rowcolor[HTML]{E5F6FF} 
\multirow{-2}{*}{\cellcolor[HTML]{E5F6FF}Data Tools} & 15 & LongDocRAG & Using this action to construct a retrieval knowledge base from the uploaded files and query the information from the knowledge base. The action can only be taken when the upload files are not None & \faFont \ \faFileTextO  \\
\bottomrule
\end{tabular}%
}
\caption{The description of the tools in the clinical toolbox. The column of Input shows the tools' input format, which indicates the form of the information that the tool can leverage. Specifically, `\faFont' indicates free text, `\faFilePhotoO' indicates the medical image, `\faFileCodeO' indicates the Python or SQL command, and `\faFileTextO' indicates the documentation file. Input with two icons indicates the tools need two types of inputs.}
\label{tab: toolbox}
\end{table*}

\subsubsection{Numerical Analysis}
Numerical analysis mainly contains two perspectives. One is the numerical calculation, commonly used in test results analysis and risk prediction~\citep{jin2024agentmd}.
We choose MedCalc~\citep{khandekar2024medcalc} to evaluate the agent capacity in equation-based and rule-based calculation tasks.
The other is the database operation and understanding, where agents need to gather patient information from the EHR database to conduct further analysis. We evaluate agents on EHRSQL~\citep{lee2022ehrsql}, MIMIC-III~\citep{johnson2016mimic}, eICU~\citep{pollard2018eicu}.
\paragraph{MedCalc} MedCalc~\citep{khandekar2024medcalc} is a novel dataset designed to evaluate the medical calculation capabilities of large language models (LLMs). It contains over 1,000 manually reviewed instances spanning 55 different medical calculation tasks, including both rule-based and equation-based calculations. Each instance provides a patient note, a specific medical question, a ground truth answer, and a step-by-step explanation. The dataset aims to assess LLMs' ability to handle medical calculations commonly used in clinical settings, focusing on arithmetic computations and extraction of relevant attributes from patient notes.

\paragraph{EHRSQL} EHRSQL~\citep{lee2022ehrsql} is a practical text-to-SQL benchmark designed for electronic health records (EHRs). It consists of 24,411 natural questions collected from 222 hospital staff, including physicians, nurses, and administrators, aimed at addressing various data retrieval needs from EHR databases. The dataset includes both answerable and unanswerable questions to evaluate trustworthy QA systems that can refuse unanswerable queries. Specifically, we only keep the answerable question in the dataset and putMed the unanswerable parts into the EHR-Halt dataset in the Trustworthiness capacity dimension. 

\paragraph{MIMIC-III}
MIMIC-III~\citep{johnson2016mimic}\footnote{\url{https://physionet.org/content/mimiciii/1.4}} covers 38,597 patients and 49,785 hospital admissions information in critical care units at the Beth Israel Deaconess Medical Center ranging from 2001 to 2012. It includes deidentified administrative information such as demographics and highly granular clinical information, including vital signs, laboratory results, procedures, medications, caregiver notes, imaging reports, and mortality. Our datasets are derived from the code base of \citet{lee2022ehrsql}. Specifically, we only keep the answerable question in the dataset and put the unanswerable parts into the EHR-Halt dataset in the Trustworthiness capacity dimension. 

\paragraph{eICU}
Similar to MIMIC-III, eICU~\citep{pollard2018eicu}\footnote{\url{https://physionet.org/content/eicu-crd/2.0}} includes over 200,000 admissions from multiple critical care units across the United States in 2014 and 2015. It contains unidentified administrative information following the US Health Insurance Portability and Accountability Act (HIPAA) standard and structured clinical data, including vital signs, laboratory measurements, medications, treatment plans, admission diagnoses, and medical histories. We also derive the dataset from the code base of \citet{lee2022ehrsql}. Specifically, we only keep the answerable question in the dataset and put the unanswerable parts into the EHR-Halt dataset in the Trustworthiness capacity dimension. 

\subsubsection{Data Understanding}
Clinical data understanding is the focus of conventional
medical natural language processing~(NLP). Agents are required to extract the key information or relations from the redundant report to understand the patient stats better. Therefore, we choose the name entity recognition dataset MedMentions~\citep{mohan2019medmentions}, information extraction dataset emrQA~\citep{pampari2018emrqa}, and long context QA dataset LongHealthQA~\citep{adams2024longhealth} to evaluate the data understanding capacity of the agents.
\paragraph{MedMentions} MedMentions~\citep{mohan2019medmentions} is a biomedical corpus consisting of over 4,000 abstracts sourced from PubMed and manually annotated with over 350,000 linked mentions of concepts from the Unified Medical Language System (UMLS). It covers a wide range of biomedical disciplines and includes over 3 million concepts from the UMLS 2017 release. MedMentions aims to support research in biomedical named entity recognition and entity linking, providing a rich resource for developing systems with broad coverage of biomedical concepts.

\paragraph{emrQA}
emrQA~\citep{pampari2018emrqa} is a large-scale question-answering (QA) dataset specifically designed for clinical notes. It is constructed by leveraging existing expert annotations from the i2b2\footnote{\url{https://www.i2b2.org/NLP/DataSets/}} datasets, resulting in a dataset with over 1 million question-logical form pairs and more than 400,000 question-answer evidence pairs. emrQA aims to support the development of QA systems capable of understanding complex clinical narratives and providing answers based on longitudinal patient records. Here, we derive the emrQA dataset from Huggingface\footnote{\url{https://huggingface.co/datasets/Eladio/emrqa-msquad}}.

\paragraph{LongHealthQA}
LongHealthQA~\citep{adams2024longhealth} is a comprehensive benchmark designed to evaluate the capabilities of LLMs in processing and interpreting extensive clinical documentation. This benchmark consists of 20 detailed fictional patient cases across various diseases, with each case containing between 5,090 to 6,754 words. The LongHealthQA benchmark challenges LLMs with 400 multiple-choice questions categorized into information extraction, negation, and sorting, providing a robust assessment tool for LLMs in the healthcare context. In this paper, we simulate multiple documentation question-answering scenarios by randomly selecting numeral other cases and constructing the LongHealthQA with context longer than 22k tokens. There are 400 questions in LongHealthQA; we chose 200 as the optimization samples and the other as the test samples.

\subsubsection{Trustworthiness}
For the application of the clinical agents, the trustworthiness of the response is very important. If clinical agents experience hallucinations while completing tasks, their responses may result in severe medical accidents. To comprehensively evaluate the hallucination that happened in the agents' solving process, we choose four types of datasets to validate the trustworthiness in four types of tasks: MedHalt-Rht~\citep{Medhalt}, MedVQA-Halt~\citep{wu2024hallucination}, EHR-Halt~\citep{lee2022ehrsql,johnson2016mimic,pollard2018eicu}, and LongHalt~\citep{adams2024longhealth}. 
\paragraph{MedHalt-Rht} Med-HALT (Medical Domain Hallucination Test) is a comprehensive benchmark and dataset for evaluating hallucination in large language models (LLMs) within the medical domain. It includes reasoning and memory-based hallucination tests, with data derived from multinational medical examinations such as AIIMS (India), USMLE (U.S.), and more. Med-HALT aims to improve the safety and reliability of LLMs in healthcare by evaluating their problem-solving and information retrieval abilities under scenarios that could induce hallucinations.

\paragraph{MedVQA-Halt} This benchmark is designed to evaluate hallucination in medical visual question answering (Med-VQA) using medical images paired with question-answer sets. It aims to assess state-of-the-art large vision and language models' performance in detecting and avoiding hallucinatory responses. The benchmark includes modified versions of existing VQA datasets like PMC-VQA, PathVQA, and VQA-RAD, with scenarios such as fake questions, "None of the Above" (NOTA), and image swaps to test the models' robustness against hallucination.

\paragraph{EHR-Halt} EHR-Halt is the trustworthiness dataset with SQL database question-answering. The dataset is constructed with the unanswerable questions derived from the EHRSQL, MIMIC-III, and eICU, resulting in 1032 samples. For this type of question, the agents need to generate the correct SQL command and retrieve it with the blank value.

\paragraph{LongHalt} Similarly to EHR-Halt, LongHalt is also derived from LongHealthQA after the operation described in \citet{adams2024longhealth}. We randomly sample multiple documentation except for the note containing the answer to the question. The model without the answer in the context can only refuse to answer the question.

\subsection{Clinical Toolbox}
\label{appendix: toolbox}
In this section, we introduce the details of the implementation of each tool in the clinical toolbox proposed. The description of the tool in the pre-built toolbox is shown in Table~\ref{tab: toolbox}.
\subsubsection{Knowledge Tools}
\paragraph{Google Search} We use the implementation of the python library \texttt{googlesearch-python}\footnote{\url{https://pypi.org/project/googlesearch-python/}}. To avoid the context becoming too lengthy due to retrieval results, we have only used the titles and abstracts of the first ten retrieved results.

\paragraph{Medrag} Following the implementation of \citet{xiong-etal-2024-benchmarking}, the knowledge base in our method is composed of PubMed\footnote{\url{https://pubmed.ncbi.nlm.nih.gov/}}, StatPearls\footnote{\url{https://www.statpearls.com/}}, and Textbooks~\citep{jin2021disease}. We adopt BM25\footnote{\url{https://github.com/facebookresearch/faiss}}~\citep{robertson2009probabilistic} as the retriever to search the information from the medical knowledge base.

\paragraph{DrugBank} We download the drug table from the website of \texttt{DrugBankOnline}\footnote{\url{https://go.drugbank.com/}}. The table consists of drugs and their relative information, including description, state, indication, dosages, and so on.

\paragraph{UMLS} For UMLS knowledge graph, we use the API provided by the National Institute of Health~(NIH)\footnote{\url{https://www.nlm.nih.gov/research/umls/index.html}}.

\subsubsection{MultiModal Tools}
\paragraph{HuatuoGPT} HuatuoGPT is a medical MLLM for medical image understanding. We adopt \texttt{HuatuoGPT-Vision-7B}\footnote{\url{https://huggingface.co/FreedomIntelligence/HuatuoGPT-Vision-7B}} as the medical image information gather. HuatuoGPT will provide the answer to the question generated by the agents.

\paragraph{MedCaptioner} MedCaptioner~\citep{xie2024medtrinity25mlargescalemultimodaldataset} is a medical image captioner which can generate the caption without any query. Besides, it decomposes the image report into five parts, including the Modality Classification, Structure Detection, ROI Analysis, Leison Texture, and Local-global Relation. The structured report gives a comprehensive description and can help the agent to better understand the medical image.

\paragraph{MiniCPM} MiniCPM~\citep{hu2024minicpm} is the MLLM developed for general domains. Here, we choose general domain MLLM as a supplement to medical MLLM to provide more options for the agent and improve the robustness of image understanding.

\subsubsection{Numerical Tools}
\paragraph{Calculator} The calculator is used to receive mathematical expressions generated by the agent and execute the calculations using Python logic. This functionality is implemented with Python’s built-in \texttt{eval} function.

\paragraph{DBManual} For DBManual, we refer to the implementation of the SQL knowledge base in EHRAgent~\citep{shi2024ehragent}, providing a detailed description of each SQL database, including its tables and columns. This allows the model to utilize DBManual to understand the structure of each database and the semantics of its columns, thereby improving task performance.

\paragraph{SQLCoder} The role of SQL Coder is to convert the agent-generated intent into SQL queries and return the results retrieved from the database. In the implementation, the SQL Coder and the agent share the same model. Since the model’s conversion to SQL syntax is not always successful, the SQL Coder can make up to three attempts based on the error messages encountered.

\subsubsection{Data Tools}
\paragraph{Spacy} SpaCy is used to extract all medical-specific terms from the input paragraph. To enhance the efficiency of entity extraction, we employ the \texttt{en\_core\_sci\_sm}\footnote{\url{https://allenai.github.io/scispacy/}} model from SciSpaCy\footnote{\url{https://github.com/allenai/scispacy}} as the NER model.

\paragraph{LongDocRAG} LongDocRAG is utilized to divide multiple user-uploaded documents into chunks and perform retrieval using Retrieval-Augmented Generation (RAG), enabling the agent to handle long contexts. In the implementation, we employ Llama-Index\footnote{\url{https://www.llamaindex.ai/}} to accomplish this operation. It is worth noting that Llama-Index can handle multimodal data; however, in this work, we limit its use to processing textual data only.

\subsection{Further Discussion}
\subsubsection{Clinical Tool}
For the coverage of tools in the proposed Clinical Toolbox, it is intractable to account for any existing medical tool due to the diversity and complexity of medical tasks. However, to ensure that our work can be broadly applicable across various clinical scenarios, we have made a conscious effort to include a diverse range of tools. For \textbf{Knowledge \& Reasoning} dimension, where task types are relatively uniform, we constructed a set of heterogeneous knowledge tools, including free text (Medrag), knowledge graphs (UMLS), tables (DrugBank), and search engines (Google Search). For the MultiModal dimension, three 
On the other hand, dimensions with inherently diverse task types naturally feature a wider variety of tools. Furthermore, our proposed method, ReflecTool, imposes no restrictions on tool types, enabling it to generalize effectively to other tools. In terms of \textbf{MultiModal} dimensions, the three selected multimodal models exhibit distinct areas of specialization. HuatuoGPT is tailored to the medical domain, providing domain-specific capabilities. MiniCPM serves as a general-purpose model, supplementing the medical model to enhance the diversity and functionality of the tools. MedCaptioner, designed specifically for generating descriptions of medical images, operates independently of queries provided by agents, thereby offering a unique utility within the framework. For other dimensions defined by diversity, a variety of tools is naturally required to ensure the comprehensive diversity of tools within the Clinical Toolbox.

\subsubsection{Clinical Scenarios}
Following the discussion on the comprehensiveness of tools, this section examines the medical scenarios in which agents can be applied. The five proposed dimensions correspond to five distinct types of scenarios anticipated for agent deployment. The \textbf{Knowledge \& Reasoning} dimension equips the model with the ability to leverage medical knowledge for tasks such as medical question-answering and reasoning-based decision-making in clinical diagnoses. The \textbf{MultiModal} dimension enables the model to process diverse types of medical images. For instance, the selected multimodal medical models, HuatuoGPT and MedCaption, support the analysis of all common imaging modalities, including X-rays, CT scans, and MRI, thereby ensuring robust coverage in medical image interpretation. The \textbf{Numerical Analysis} dimension allows the model to integrate with hospital Electronic Health Record (EHR) systems, enhancing its capacity to assist physicians in delivering medical services. This capability also improves the model’s proficiency in interpreting numerical data, enabling accurate evaluations of clinical indicators for normalcy. The \textbf{Data Understanding} dimension enhances the model’s ability to process medical text, such as patient reports and long-form records like longitudinal follow-ups. Lastly, the \textbf{Trustworthiness} dimension focuses on minimizing hallucinations during medical task execution, thereby improving the model’s reliability in clinical applications.


\section{Case Study}
\subsection{Parameter Error in Tool Usage}
\label{appendix: parameter error}
\begin{figure*}[tbp]
    \centering
    \includegraphics[width=1.0\linewidth]{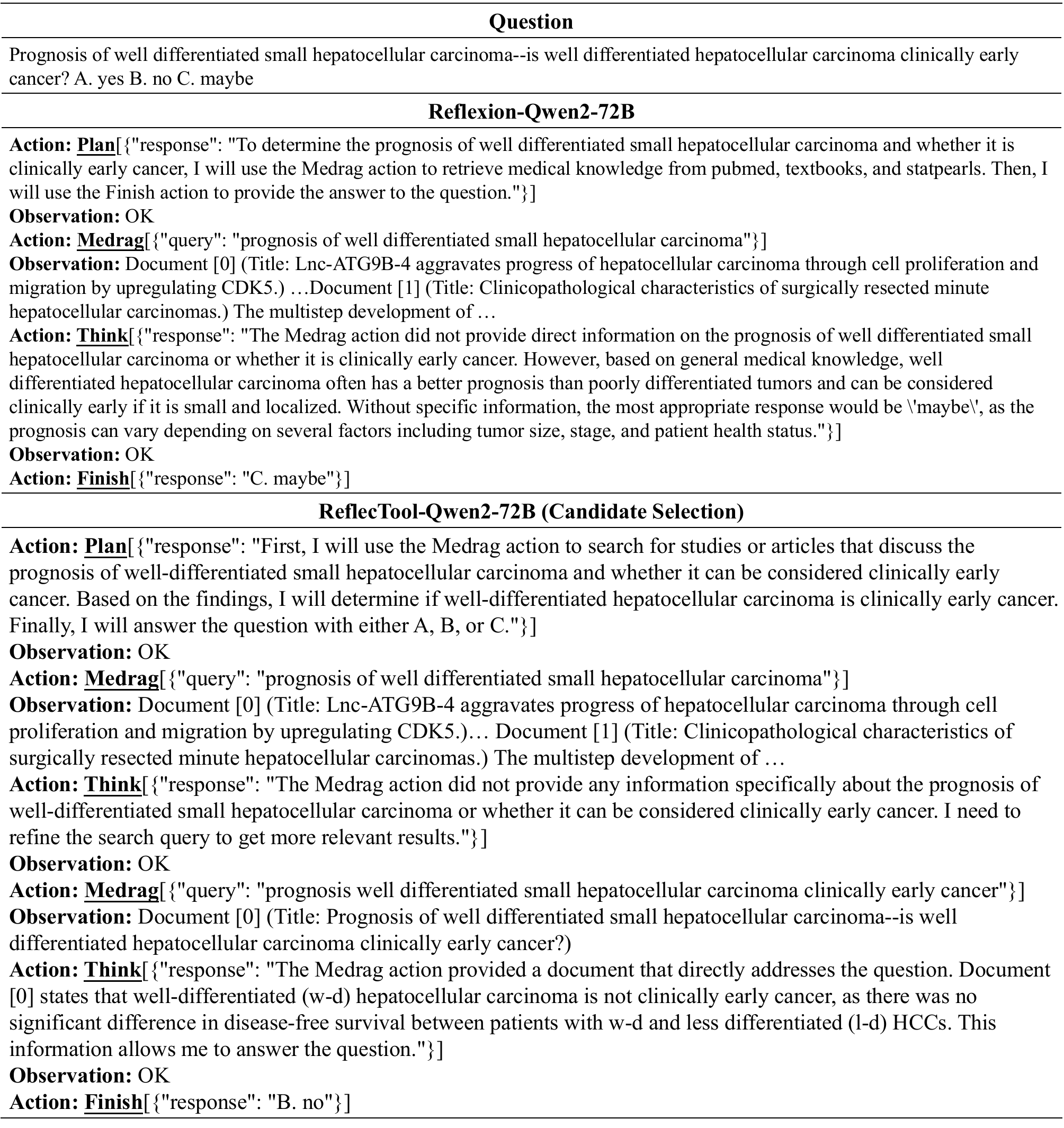}
    \caption{Case study for parameter error in tool usage. The figure compares the reasoning trajectories of two agent-based methods for the same task. }
    \label{fig: case study param error}
\end{figure*}

There are two types of errors in the process of an agent utilizing tools to solve problems. The first is tool selection errors, as discussed in Section~\ref{sec: tool error}, where the agent employs an inappropriate tool for the task. The second is parameter errors, which occur when the agent uses the correct tool but fails to provide appropriate parameters, leading to unexpected results. A common example is when an agent invokes a knowledge tool but generates an unsuitable query, causing the tool to retrieve knowledge with low relevance to the task, thereby failing to provide the agent with effective information. Figure~\ref{fig: case study param error} presents a case study to illustrate this phenomenon more clearly. It can be observed that while both agents invoked the same tool, Medrag, Reflexion-Qwen2-72B failed to obtain the correct result. In contrast, ReflecTool-Qwen2-72B, despite initially failing to retrieve relevant information, successfully generated the correct query after reflection, ultimately retrieving effective information. This demonstrates that ReflecTool not only enhances the model’s ability to select the appropriate domain-specific tools but also enables the agent to leverage the tool-wise reflection mechanism to provide better parameters when invoking tools, thereby maximizing their effectiveness.

\section{Experiments Details}
\label{sec: experiments details}
\subsection{Implementations}
All the LLMs and MLLMs methods solve the task under the \textit{zero-shot} settings while all the agent-based methods are \textit{one-shot} to demonstrate the action format for the agents. Besides, considering that the multimodal tools can only be used in multimodal tasks, we remove the multimodal tools when agents are solving other types of tasks to save the cost of memory. All the experiments are run on two NVIDIA A100 80GB.

\subsection{Prompt Used in \agentname}
\begin{figure*}[tbp]
    \centering
    \includegraphics[width=1.0\linewidth]{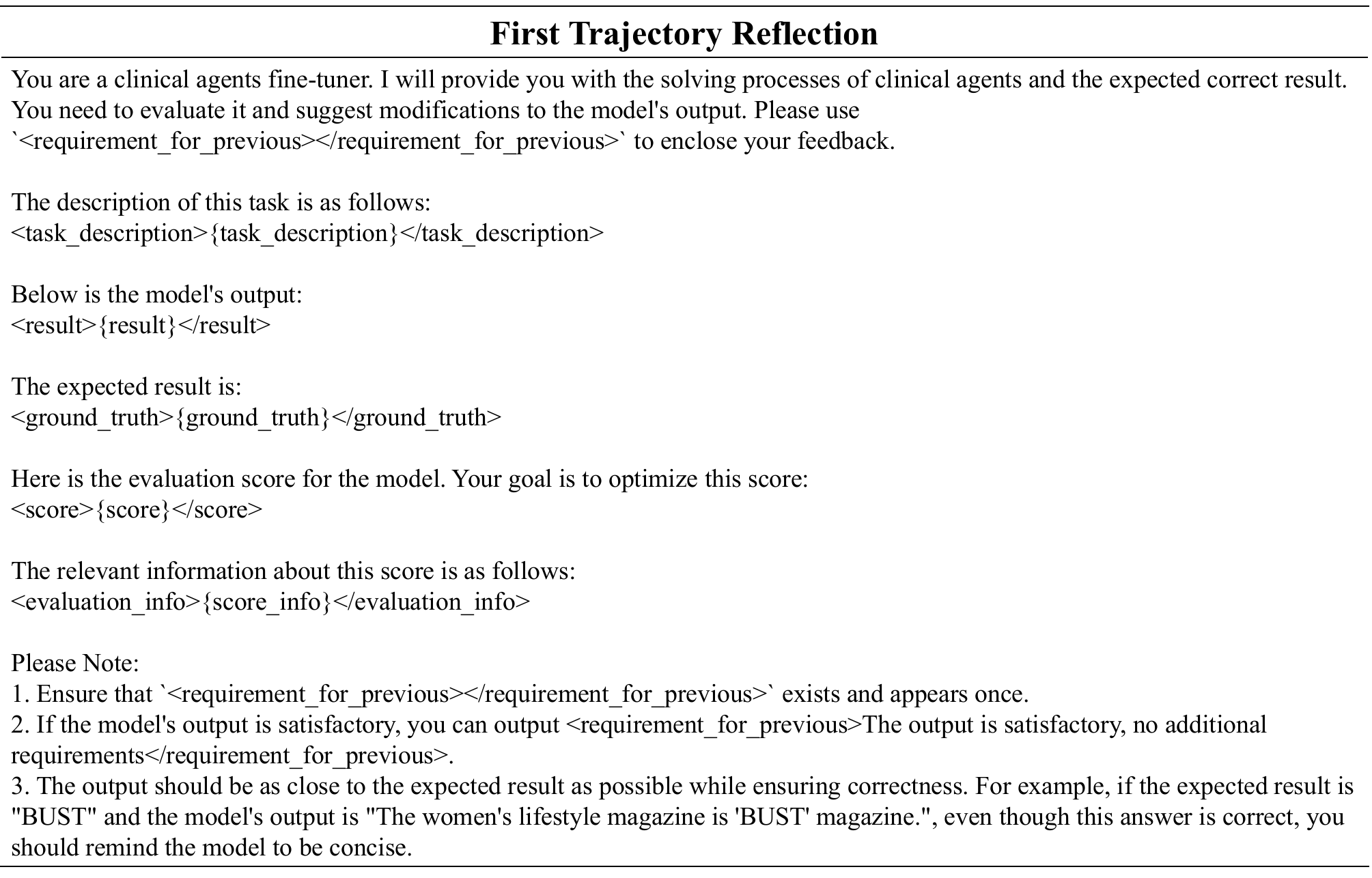}
    \caption{Prompt for the reflection on the first trajectory in optimization stage. }
    \label{fig: reflect prompt}
\end{figure*}

\begin{figure*}[tbp]
    \centering
    \includegraphics[width=1.0\linewidth]{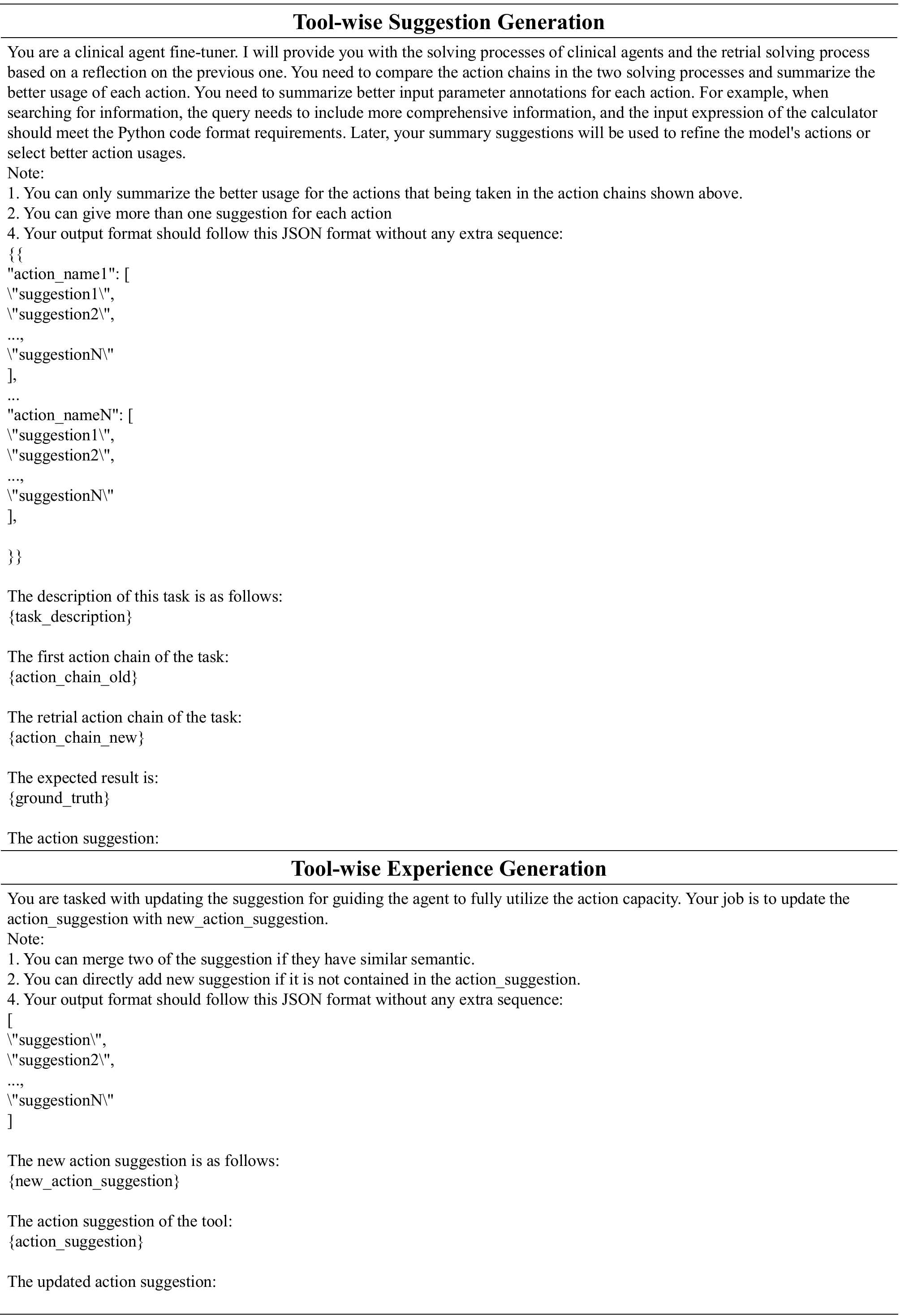}
    \caption{Prompt for the tool-wise suggestion and tool-wise experience generation.}
    \label{fig: tool-wise prompt}
\end{figure*}

\begin{figure*}[t]
    \centering
    \includegraphics[width=1.0\linewidth]{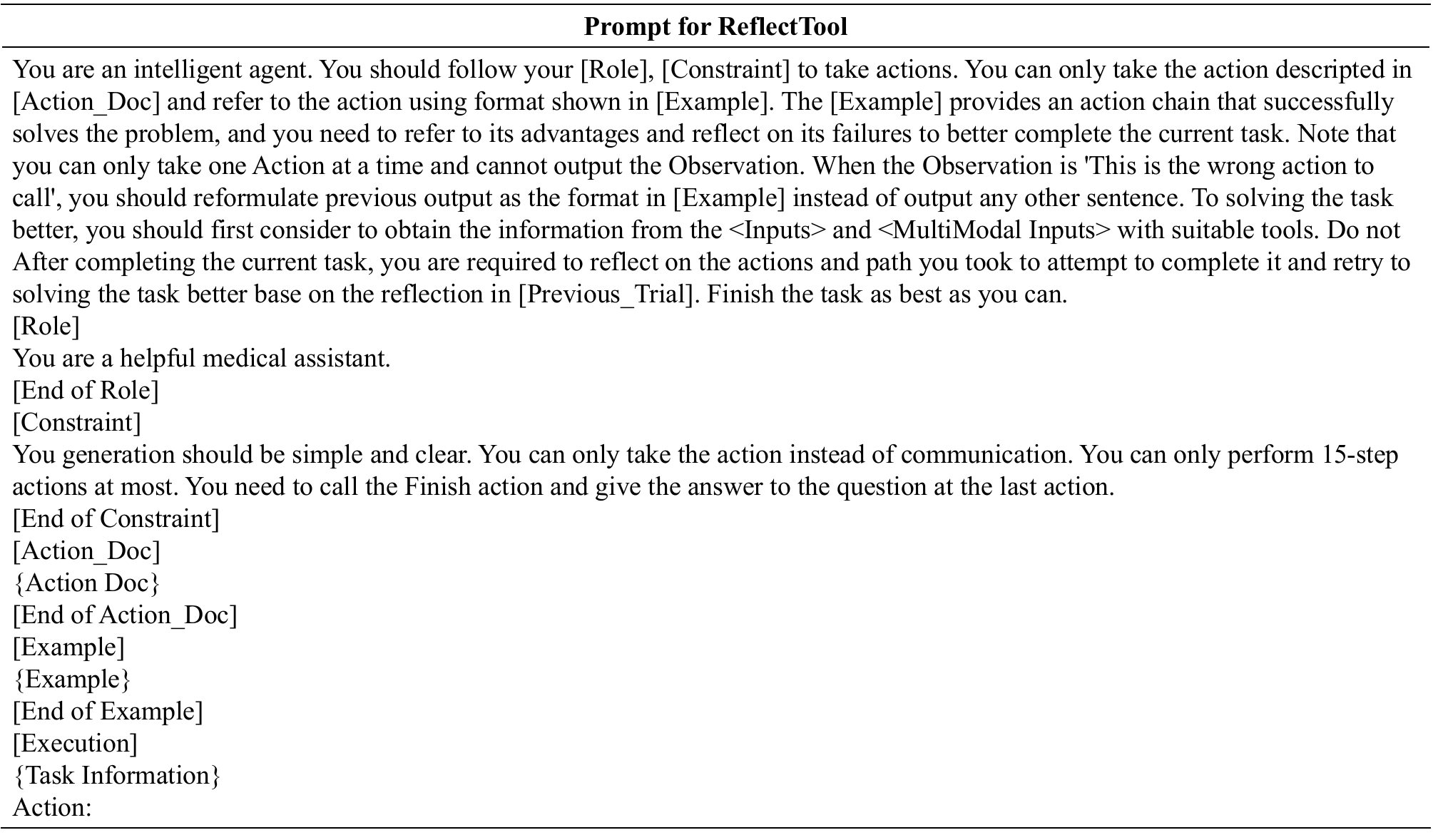}
    \caption{Prompt for the \agentname. }
    \label{fig: prompt}
\end{figure*}

\begin{figure*}[t]
    \centering
    \includegraphics[width=1.0\linewidth]{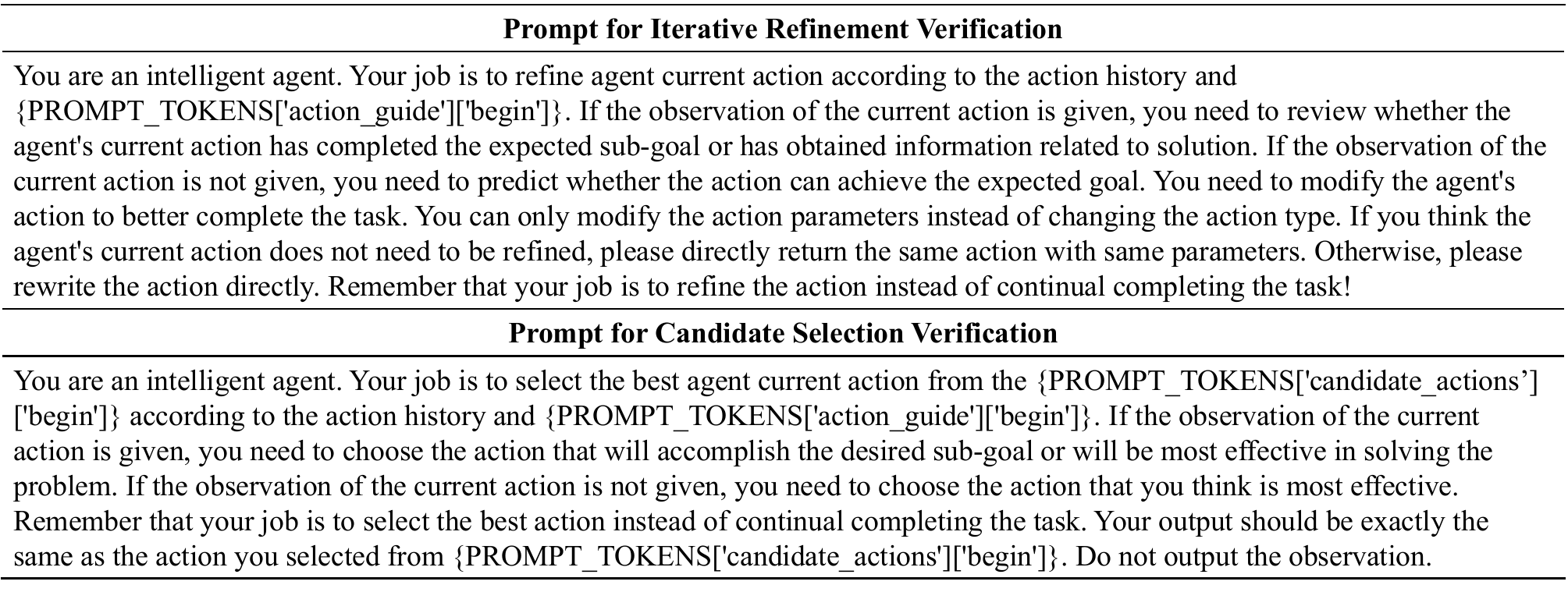}
    \caption{Prompt for two type of Verifier. }
    \label{fig: verification prompt}
\end{figure*}

The prompts of the optimization stage are shown in Figure~\ref{fig: reflect prompt} and Figure~\ref{fig: tool-wise prompt}. The prompts of the inference stage are shown in Figure~\ref{fig: prompt}, and two types of verifiers are shown in Figure~\ref{fig: verification prompt}.

\begin{landscape}
\begin{table}[tbp]
\renewcommand\arraystretch{1.2}
\centering
\resizebox{\linewidth}{!}{%
\begin{tabular}{lcccccccccccccccccccccccc}
\toprule
 & \multicolumn{5}{c}{\textbf{Knowledge\&Reasoning}} & \multicolumn{4}{c}{\textbf{MultModal}} & \multicolumn{5}{c}{\textbf{Numerical Analysis}} & \multicolumn{4}{c}{\textbf{Data Understanding}} & \multicolumn{5}{c}{\textbf{Trustworthiness}} & \cellcolor[HTML]{F2F2F2} \\
 \cmidrule(lr){2-6}\cmidrule(lr){7-10}\cmidrule(lr){11-15}\cmidrule(lr){16-19}\cmidrule(lr){20-24}
\multirow{-2}{*}{\textbf{Methods}} & \textbf{MedQA} & \textbf{MMLU} & \textbf{BioASQ} & \textbf{\begin{tabular}[c]{@{}c@{}}Pub\\ MedQA\end{tabular}} & \cellcolor[HTML]{FFE9E8}\textbf{Avg.} & \textbf{\begin{tabular}[c]{@{}c@{}}VQA\\ RAD\end{tabular}} & \textbf{SLAKE} & \textbf{\begin{tabular}[c]{@{}c@{}}Omni\\ MedQA\end{tabular}} & \cellcolor[HTML]{FFF9E3}\textbf{Avg.} & \textbf{MedCalc} & \textbf{\begin{tabular}[c]{@{}c@{}}EHR\\ SQL\end{tabular}} & \textbf{\begin{tabular}[c]{@{}c@{}}MIMIC\\ -III\end{tabular}} & \textbf{eICU} & \cellcolor[HTML]{EAFAF1}\textbf{Avg.} & \textbf{\begin{tabular}[c]{@{}c@{}}Med\\ Mentions\end{tabular}} & \textbf{emrQA} & \textbf{\begin{tabular}[c]{@{}c@{}}Long\\ HealthQA\end{tabular}} & \cellcolor[HTML]{E5F6FF}\textbf{Avg.} & \textbf{\begin{tabular}[c]{@{}c@{}}MedHalt\\ -Rht\end{tabular}} & \textbf{\begin{tabular}[c]{@{}c@{}}MedVQA\\ -Halt\end{tabular}} & \textbf{\begin{tabular}[c]{@{}c@{}}EHR\\ -Halt\end{tabular}} & \textbf{\begin{tabular}[c]{@{}c@{}}Long\\ Halt\end{tabular}} & \cellcolor[HTML]{FDEBFF}\textbf{Avg.} & \multirow{-2}{*}{\cellcolor[HTML]{F2F2F2}\textbf{Total}} \\
\midrule
\multicolumn{25}{c}{\textit{Large Language Models}} \\
\midrule
MedLlama3-8b & 58.13 & 72.82 & 66.18 & 42.40 & \cellcolor[HTML]{FFE9E8}59.88 & - & - & - & \cellcolor[HTML]{FFF9E3}- & 22.45 & 7.92 & 8.65 & 7.40 & \cellcolor[HTML]{EAFAF1}11.61 & 22.93 & 23.10 & - & \cellcolor[HTML]{E5F6FF}23.02 & 9.90 & - & 2.23 & - & \cellcolor[HTML]{FDEBFF}6.07 & \cellcolor[HTML]{F2F2F2}25.14 \\
Qwen2-7B~\citep{qwen2} & 54.04 & 69.51 & 71.68 & 48.20 & \cellcolor[HTML]{FFE9E8}60.86 & - & - & - & \cellcolor[HTML]{FFF9E3}- & 14.42 & 18.09 & 19.70 & 21.73 & \cellcolor[HTML]{EAFAF1}18.49 & 18.38 & 39.86 & 75.25 & \cellcolor[HTML]{E5F6FF}44.50 & 14.11 & - & 3.97 & 66.50 & \cellcolor[HTML]{FDEBFF}28.19 & \cellcolor[HTML]{F2F2F2}38.01 \\
Llama3-8b~\citep{llama3modelcard} & 56.32 & 70.34 & 72.82 & 53.80 & \cellcolor[HTML]{FFE9E8}63.32 & - & - & - & \cellcolor[HTML]{FFF9E3}- & 28.08 & 17.02 & 12.53 & 21.83 & \cellcolor[HTML]{EAFAF1}19.87 & 28.54 & 41.92 & - & \cellcolor[HTML]{E5F6FF}35.23 & 29.22 & - & 18.90 & - & \cellcolor[HTML]{FDEBFF}24.06 & \cellcolor[HTML]{F2F2F2}35.62 \\
Llama3.1-8b~\citep{DBLP:journals/corr/abs-2407-21783} & 65.20 & 76.58 & 74.92 & 53.00 & \cellcolor[HTML]{FFE9E8}67.43 & - & - & - & \cellcolor[HTML]{FFF9E3}- & 37.44 & 11.35 & 17.42 &  & \cellcolor[HTML]{EAFAF1}22.07 & 32.08 & 42.41 & 74.25 & \cellcolor[HTML]{E5F6FF}49.58 & 27.78 & - & 3.97 & 60.50 & \cellcolor[HTML]{FDEBFF}30.75 & \cellcolor[HTML]{F2F2F2}42.46 \\
Qwen2-72B*~\citep{qwen2} & 71.25 & 84.48 & 82.85 & 53.00 & \cellcolor[HTML]{FFE9E8}72.90 & - & - & - & \cellcolor[HTML]{FFF9E3}- & 32.19 & 23.98 & 33.95 & 34.15 & \cellcolor[HTML]{EAFAF1}31.07 & 29.20 & 42.89 & 79.75 & \cellcolor[HTML]{E5F6FF}50.61 & 31.56 & - & 31.30 & 58.50 & \cellcolor[HTML]{FDEBFF}40.45 & \cellcolor[HTML]{F2F2F2}48.76 \\
Llama3.1-70b*~\citep{DBLP:journals/corr/abs-2407-21783} & 79.58 & 88.15 & 82.52 & 57.40 & \cellcolor[HTML]{FFE9E8}\textbf{76.91} & - & - & - & \cellcolor[HTML]{FFF9E3}- & 48.52 & 16.49 & 25.44 & 26.47 & \cellcolor[HTML]{EAFAF1}29.23 & 25.71 & 31.69 & 80.00 & \cellcolor[HTML]{E5F6FF}45.80 & 28.22 & - & 22.48 & 64.50 & \cellcolor[HTML]{FDEBFF}38.40 & \cellcolor[HTML]{F2F2F2}47.59 \\
GPT-3.5-turbo~\citep{elmohamed} & 58.68 & 69.88 & 75.40 & 50.60 & \cellcolor[HTML]{FFE9E8}63.64 & - & - & - & \cellcolor[HTML]{FFF9E3}- & 20.53 & 17.57 & 24.31 & 14.30 & \cellcolor[HTML]{EAFAF1}19.18 & 26.88 & 21.64 & - & \cellcolor[HTML]{E5F6FF}24.26 & 9.78 & - & 26.55 & - & \cellcolor[HTML]{FDEBFF}18.17 & \cellcolor[HTML]{F2F2F2}31.31 \\
\midrule
\multicolumn{25}{c}{\textit{MultiModal Large Language Models}} \\
\midrule
MiniCPM-V-2.6~\citep{yao2024minicpm} & 46.58 & 61.16 & 70.23 & 47.20 & \cellcolor[HTML]{FFE9E8}56.29 & 48.78 & 47.12 & 73.70 & \cellcolor[HTML]{FFF9E3}56.53 & 13.28 & 1.61 & 1.63 & 1.88 & \cellcolor[HTML]{EAFAF1}4.60 & 18.92 & 17.42 & 5.25 & \cellcolor[HTML]{E5F6FF}13.86 & 12.44 & 36.89 & 8.91 & 2.25 & \cellcolor[HTML]{FDEBFF}15.12 & \cellcolor[HTML]{F2F2F2}29.28 \\
InternVL-Chat-V1.5~\citep{chen2023internvl} & 50.82 & 65.56 & 64.89 & 30.40 & \cellcolor[HTML]{FFE9E8}52.92 & 49.67 & 41.47 & 68.50 & \cellcolor[HTML]{FFF9E3}53.21 & 18.91 & 17.99 & 18.05 & 20.83 & \cellcolor[HTML]{EAFAF1}18.95 & 26.47 & 42.53 & - & \cellcolor[HTML]{E5F6FF}34.50 & 25.78 & 50.78 & 0.00 & * & \cellcolor[HTML]{FDEBFF}25.52 & \cellcolor[HTML]{F2F2F2}37.02 \\
HuatuoGPT-Vision-7B~\citep{DBLP:journals/corr/abs-2406-19280} & 50.43 & 66.12 & 73.30 & 54.00 & \cellcolor[HTML]{FFE9E8}60.96 & 53.65 & 52.97 & 92.10 & \cellcolor[HTML]{FFF9E3}66.24 & 13.56 & 4.39 & 9.27 & 9.76 & \cellcolor[HTML]{EAFAF1}9.25 & 16.74 & 38.44 & 73.00 & \cellcolor[HTML]{E5F6FF}42.73 & 14.33 & 23.44 & 2.42 & 65.25 & \cellcolor[HTML]{FDEBFF}26.36 & \cellcolor[HTML]{F2F2F2}41.11 \\
HuatuoGPT-Vision-34B~\citep{DBLP:journals/corr/abs-2406-19280} & 54.83 & 72.36 & 73.79 & 48.00 & \cellcolor[HTML]{FFE9E8}62.25 & 56.76 & 53.72 & 91.50 & \cellcolor[HTML]{FFF9E3}\textbf{67.33} & 25.79 & 8.14 & 9.15 & 9.79 & \cellcolor[HTML]{EAFAF1}13.22 & 16.34 & 41.68 & - & \cellcolor[HTML]{E5F6FF}29.01 & 28.44 & 43.77 & 32.07 & * & \cellcolor[HTML]{FDEBFF}34.76 & \cellcolor[HTML]{F2F2F2}41.31 \\
GPT-4o-mini~\citep{gpt4} & 76.90 & 85.67 & 82.84 & 49.20 & \cellcolor[HTML]{FFE9E8}73.65 & 50.47 & 46.47 & 59.20 & \cellcolor[HTML]{FFF9E3}52.05 & 50.43 & 21.73 & 28.07 & 18.19 & \cellcolor[HTML]{EAFAF1}29.61 & 31.66 & 40.00 & 78.50 & \cellcolor[HTML]{E5F6FF}50.05 & 62.11 & 53.78 & 45.74 & 70.00 & \cellcolor[HTML]{FDEBFF}\textbf{57.91} & \cellcolor[HTML]{F2F2F2}52.65 \\
\midrule
\multicolumn{25}{c}{\textit{Agent (Qwen2-7b)}} \\
\midrule
COT~\citep{DBLP:conf/nips/Wei0SBIXCLZ22} & 52.47 & 69.97 & 72.21 & 41.00 & \cellcolor[HTML]{FFE9E8}58.91 & - & - & - & \cellcolor[HTML]{FFF9E3}- & 19.10 & 16.06 & 23.81 & 20.95 & \cellcolor[HTML]{EAFAF1}19.98 & 22.83 & 19.92 & 65.75 & \cellcolor[HTML]{E5F6FF}36.17 & 45.56 & - & 31.49 & 57.00 & \cellcolor[HTML]{FDEBFF}44.68 & \cellcolor[HTML]{F2F2F2}39.94 \\
ReAct~\citep{DBLP:conf/iclr/YaoZYDSN023} & 51.61 & 67.68 & 80.24 & 48.60 & \cellcolor[HTML]{FFE9E8}62.03 & 35.92 & 39.59 & 72.90 & \cellcolor[HTML]{FFF9E3}49.47 & 18.62 & 18.52 & 25.06 & 34.00 & \cellcolor[HTML]{EAFAF1}24.05 & 22.19 & 24.04 & 41.50 & \cellcolor[HTML]{E5F6FF}29.24 & 49.89 & 35.33 & 61.49 & 48.75 & \cellcolor[HTML]{FDEBFF}53.87 & \cellcolor[HTML]{F2F2F2}42.73 \\
CRITIC~\citep{DBLP:conf/iclr/GouSGSYDC24} & 52.87 & 58.68 & 71.68 & 43.20 & \cellcolor[HTML]{FFE9E8}56.61 & 48.12 & 42.70 & 70.80 & \cellcolor[HTML]{FFF9E3}53.87 & 13.09 & 23.55 & 28.20 & 33.12 & \cellcolor[HTML]{EAFAF1}24.49 & 25.47 & 36.33 & 50.25 & \cellcolor[HTML]{E5F6FF}37.35 & 30.44 & 28.33 & 57.64 & 73.75 & \cellcolor[HTML]{FDEBFF}47.54 & \cellcolor[HTML]{F2F2F2}43.97 \\
Reflexion~\citep{DBLP:conf/nips/ShinnCGNY23} & 51.78 & 66.48 & 74.60 & 50.80 & \cellcolor[HTML]{FFE9E8}60.92 & 45.68 & 47.97 & 77.20 & \cellcolor[HTML]{FFF9E3}56.95 & 13.37 & 17.56 & 22.16 & 30.23 & \cellcolor[HTML]{EAFAF1}20.83 & 30.30 & 28.92 & 53.00 & \cellcolor[HTML]{E5F6FF}37.41 & 50.55 & 36.33 & 62.91 & 50.75 & \cellcolor[HTML]{FDEBFF}50.14 & \cellcolor[HTML]{F2F2F2}45.25 \\
\textbf{MedToolAgent (Iterative Refinement, k=2)} & 50.12 & 65.47 & 76.37 & 63.20 & \cellcolor[HTML]{FFE9E8}63.79 & 53.88 & 45.71 & 82.90 & \cellcolor[HTML]{FFF9E3}60.83 & 24.68 & 24.20 & 16.92 & 22.08 & \cellcolor[HTML]{EAFAF1}21.97 & 43.69 & 50.27 & 61.00 & \cellcolor[HTML]{E5F6FF}51.65 & 54.44 & 37.99 & 56.73 & 45.25 & \cellcolor[HTML]{FDEBFF}48.60 & \cellcolor[HTML]{F2F2F2}49.37 \\
\textbf{MedToolAgent (Candidates Selection, k=2)} & 50.81 & 64.84 & 72.98 & 62.60 & \cellcolor[HTML]{FFE9E8}62.81 & 56.76 & 49.76 & 79.20 & \cellcolor[HTML]{FFF9E3}61.91 & 24.64 & 29.87 & 25.13 & 27.48 & \cellcolor[HTML]{EAFAF1}26.78 & 59.21 & 43.40 & 54.00 & \cellcolor[HTML]{E5F6FF}52.20 & 53.44 & 34.21 & 55.97 & 23.25 & \cellcolor[HTML]{FDEBFF}41.72 & \cellcolor[HTML]{F2F2F2}49.08 \\
\midrule
\multicolumn{25}{c}{\textit{Agent (Qwen2-72b*)}} \\
\midrule
COT~\citep{DBLP:conf/nips/Wei0SBIXCLZ22} & 72.51 & 85.45 & 81.07 & 37.40 & \cellcolor[HTML]{FFE9E8}69.11 & - & - & - & \cellcolor[HTML]{FFF9E3}- & 29.89 & 18.73 & 23.55 & 25.72 & \cellcolor[HTML]{EAFAF1}24.47 & 24.43 & 52.09 & 81.00 & \cellcolor[HTML]{E5F6FF}52.51 & 57.11 & - & 40.79 & 70.75 & \cellcolor[HTML]{FDEBFF}56.22 & \cellcolor[HTML]{F2F2F2}50.58 \\
ReAct~\citep{DBLP:conf/iclr/YaoZYDSN023} & 72.43 & 85.67 & 85.38 & 62.40 & \cellcolor[HTML]{FFE9E8}76.47 & 50.09 & 46.01 & 73.00 & \cellcolor[HTML]{FFF9E3}56.37 & 26.46 & 35.97 & 31.45 & 31.87 & \cellcolor[HTML]{EAFAF1}31.44 & 42.11 & 55.02 & 62.75 & \cellcolor[HTML]{E5F6FF}53.29 & 56.89 & 24.75 & 55.78 & 58.50 & \cellcolor[HTML]{FDEBFF}48.98 & \cellcolor[HTML]{F2F2F2}53.31 \\
CRITIC~\citep{DBLP:conf/iclr/GouSGSYDC24} & 71.85 & 85.31 & 87.86 & 51.00 & \cellcolor[HTML]{FFE9E8}74.01 & 40.58 & 50.80 & 73.50 & \cellcolor[HTML]{FFF9E3}54.96 & 22.44 & 35.48 & 32.47 & 33.28 & \cellcolor[HTML]{EAFAF1}30.92 & 33.60 & 56.36 & 75.50 & \cellcolor[HTML]{E5F6FF}55.15 & 51.77 & 24.22 & 52.03 & 58.75 & \cellcolor[HTML]{FDEBFF}46.69 & \cellcolor[HTML]{F2F2F2}52.35 \\
Reflexion~\citep{DBLP:conf/nips/ShinnCGNY23} & 70.78 & 84.30 & 87.06 & 65.00 & \cellcolor[HTML]{FFE9E8}76.79 & 54.99 & 50.05 & 77.80 & \cellcolor[HTML]{FFF9E3}60.95 & 22.45 & 40.14 & 31.94 & 33.42 & \cellcolor[HTML]{EAFAF1}31.99 & 52.37 & 58.73 & 64.00 & \cellcolor[HTML]{E5F6FF}58.37 & 59.00 & 33.00 & 59.00 & 64.00 & \cellcolor[HTML]{FDEBFF}53.75 & \cellcolor[HTML]{F2F2F2}56.37 \\
\textbf{MedToolAgent (Iterative Refinement, k=2)} & 73.30 & 84.11 & 84.63 & 65.20 & \cellcolor[HTML]{FFE9E8}76.81 & 57.21 & 48.82 & 85.20 & \cellcolor[HTML]{FFF9E3}63.74 & 36.01 & 47.43 & 33.96 & 36.39 & \cellcolor[HTML]{EAFAF1}\textbf{38.45} & 54.57 & 67.96 & 68.00 & \cellcolor[HTML]{E5F6FF}63.51 & 59.55 & 38.21 & 57.82 & 63.00 & \cellcolor[HTML]{FDEBFF}54.65 & \cellcolor[HTML]{F2F2F2}59.43 \\
\textbf{MedToolAgent (Candidates Selection, k=2)} & 71.37 & 84.00 & 83.50 & 66.20 & \cellcolor[HTML]{FFE9E8}76.27 & 57.66 & 48.54 & 81.90 & \cellcolor[HTML]{FFF9E3}62.70 & 36.77 & 49.89 & 31.20 & 34.38 & \cellcolor[HTML]{EAFAF1}38.06 & 60.51 & 66.61 & 66.50 & \cellcolor[HTML]{E5F6FF}\textbf{64.54} & 60.77 & 39.63 & 59.78 & 66.75 & \cellcolor[HTML]{FDEBFF}56.73 & \cellcolor[HTML]{F2F2F2}\textbf{59.66} \\
\bottomrule
\end{tabular}%
}
\caption{Experimental results of four types of models on Clinical Agent Bench. The `COT' method indicates the agent runs without the pre-built tools. `*' indicates the models use 4-bit GPTQ quantization. `-' means the model is not capable of solving such a task. The best results of each type of task are \textbf{Bold}.}
\label{tab: main result all}
\end{table}
\end{landscape}

\end{document}